\RequirePackage{fix-cm}
\documentclass[twocolumn]{svjour3}          %
\smartqed  %
\usepackage{graphicx}
\usepackage{amsmath}
\usepackage{multirow}
\usepackage{pifont}
\usepackage{graphicx}
\usepackage{amssymb}
\usepackage{array}
\usepackage{multirow, nicefrac}
\usepackage{pifont}
\usepackage{cite}
\usepackage{bbm}
\usepackage{arydshln}
\usepackage[dvipsnames]{xcolor}
\usepackage{tablefootnote}
\usepackage{threeparttable}
\definecolor{green}{rgb}{1,0,0}
\usepackage{stfloats}
\usepackage{color}
\usepackage[normalem]{ulem}
\usepackage[switch]{lineno} 
\usepackage[symbol]{footmisc}

\def\cD{ {\cal D } }

\def\cC{\mathcal{C}}

\def\0{{\bf 0}}
\def\1{{\bf 1}}

\usepackage{booktabs}
\usepackage[utf8]{inputenc}

\DeclareRobustCommand\onedot{\futurelet\@let@token\@onedot}
\def\@onedot{\ifx\@let@token.\else.\null\fi\xspace}

\makeatother

\def\R{{\mathbb R}}
\def\atm{ATM} 

\def\seg{SegViT}

\usepackage{listings,xcolor} 
\usepackage{tcolorbox}
\newcommand{\lstfont}[1]{\color{#1}\ttfamily}

\usepackage{xspace}
\usepackage[colorlinks,linkcolor=red,anchorcolor=blue,citecolor=purple,CJKbookmarks=True]{hyperref}

\usepackage{marvosym}

\makeatletter
\def\cl@chapter{\@elt {theorem}}
\makeatother
\usepackage[capitalise]{cleveref}

\begin{document}
\sloppy
\title{SegViT v2: Exploring Efficient and 
Continual 
Semantic Segmentation with Plain Vision Transformers
}

\author{
Bowen Zhang\footnotemark[1] ,
        Liyang Liu\footnotemark[1] ,
        Minh Hieu Phan\footnotemark[1] ,
        Zhi Tian,
        Chunhua Shen,
        and Yifan Liu\footnotemark[2] %
}
\authorrunning{B. Zhang, L. Liu, M. H. Phan,  Z. Tian, C. Shen, Y. Liu} %

\institute{
Bowen Zhang, Liyang Liu, Minh Hieu Phan, Yifan Liu\at
              The University of Adelaide \\
              \email{\{b.zhang,akide.liu,vuminhhieu.phan,yifan.liu04\}\\
              @adelaide.edu.au}
           \and
           Chunhua Shen \at
           Zhejiang University \\
           \email{chunhuashen@zju.edu.cn}  \and
           Zhi Tian \at
               Meituan Inc.\\
              \email{zhi.tian}@outlook.com}

\date{\today}

\maketitle

\footnotetext[1]{These authors contributed equally to this work.}
\footnotetext[2]{Corresponding author.}

\maketitle

\begin{abstract}
This paper investigates the capability of plain Vision Transformers (ViTs) for semantic segmentation using the encoder-decoder framework and introduce \textbf{SegViTv2}. 
In this study, we introduce a novel Attention-to-Mask (\atm) module to design a lightweight decoder effective for plain ViT. The proposed ATM converts the global attention map into semantic masks for high-quality segmentation results.
Our decoder outperforms popular decoder UPerNet using various ViT backbones while consuming only about $5\%$ of the computational cost. 
For the encoder, we address the concern of the relatively high computational cost in the ViT-based encoders and propose a \emph{Shrunk++} structure that incorporates edge-aware query-based down-sampling
(EQD) and query-based up-sa\-m\-p\-l\-i\-n\-g (QU) modules.
The Shrunk++ structure reduces the computational cost of the encoder by up to $50\%$ while maintaining competitive performance.
Furthermore, we propose to adapt SegViT for continual semantic segmentation, demonstrating nearly zero forgetting of previously learned knowledge.
Experiments show that our proposed SegViTv2 surpasses recent segmentation methods on three popular benchmarks including ADE20k, COCO-Stuff-10k
and PASCAL-Context datasets.
The code is available
through the following link: 
\url{https://github.com/zbwxp/SegVit}.

\end{abstract}

\definecolor{o_mycolor}{rgb}{0.9,0.57,0.49}
\definecolor{p_mycolor}{rgb}{0.564,0.52941176,0.7254902}
\begin{figure}[ht]
    \centering
    \scalebox{0.60}{\includegraphics[width=0.8
    \textwidth]{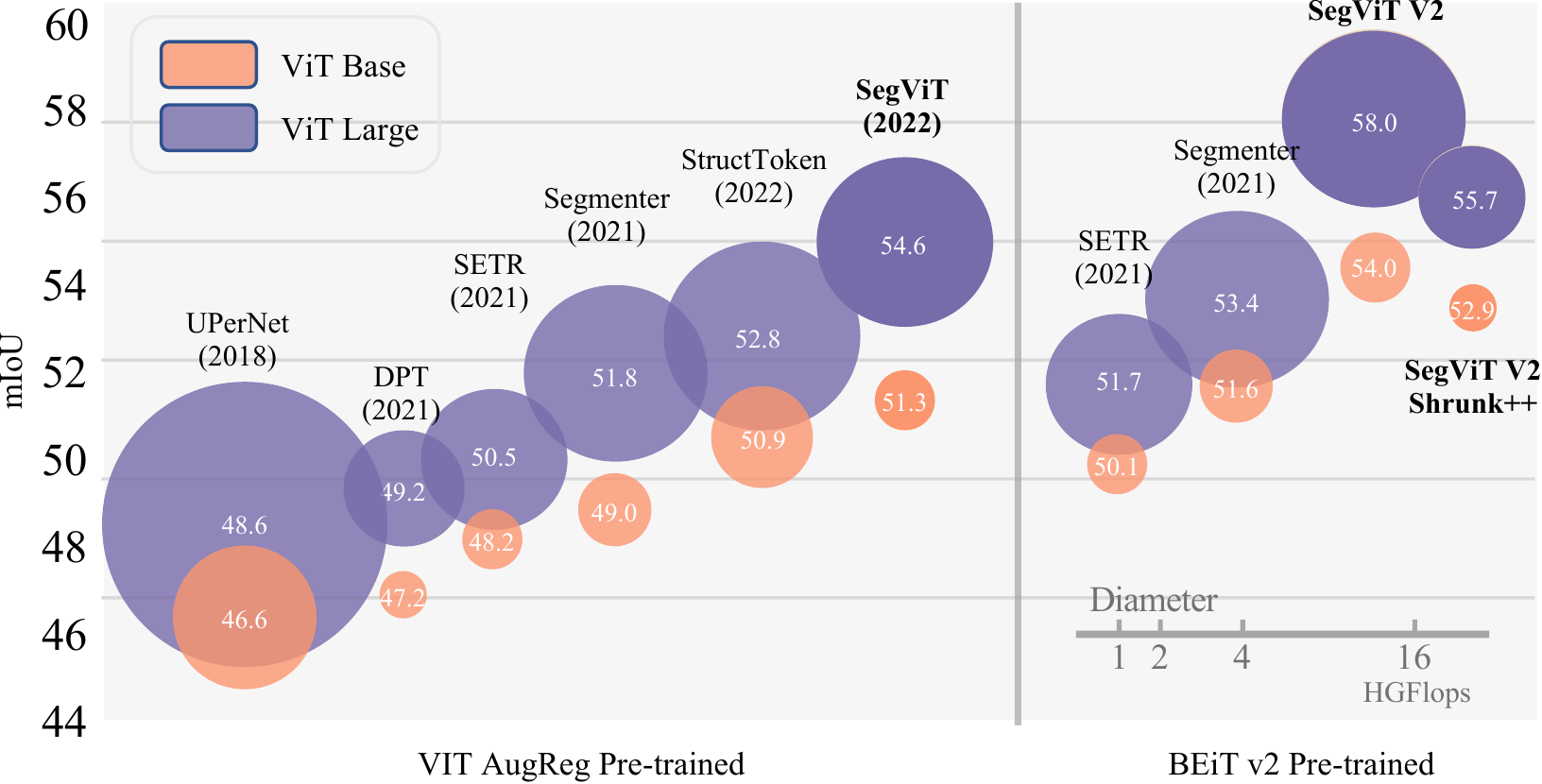}}
    \caption{\textbf{Comparison with previous methods in terms of performance and efficiency} on ADE20K dataset. The \textbf{\textcolor{o_mycolor}{orange}} and \textbf{\textcolor{p_mycolor}{purple}} bubbles in the accompanying graph represent the ViT Base and ViT Large %
    models, respectively, with the size of each bubble corresponding to the FLOPs of the variant segmentation methods. %
    SegViT-BEiT v2 Large achieves state-of-the-art performance with a \textbf{58.0\%} mIoU on the ADE20K validation set. Additionally, our efficient, optimized version, %
    SegViT-Shrunk-BEiT v2 Large, saves half of the GFLOPs compared to UPerNet, significantly reducing computational overhead while maintaining a competitive performance of \textbf{55.7\%}.
    }
\label{fig:intro_digraph_comparison}
\vspace{-6mm}
\end{figure}

\begin{figure*}[t]
    \centering
    \includegraphics[width=0.98
    \textwidth]{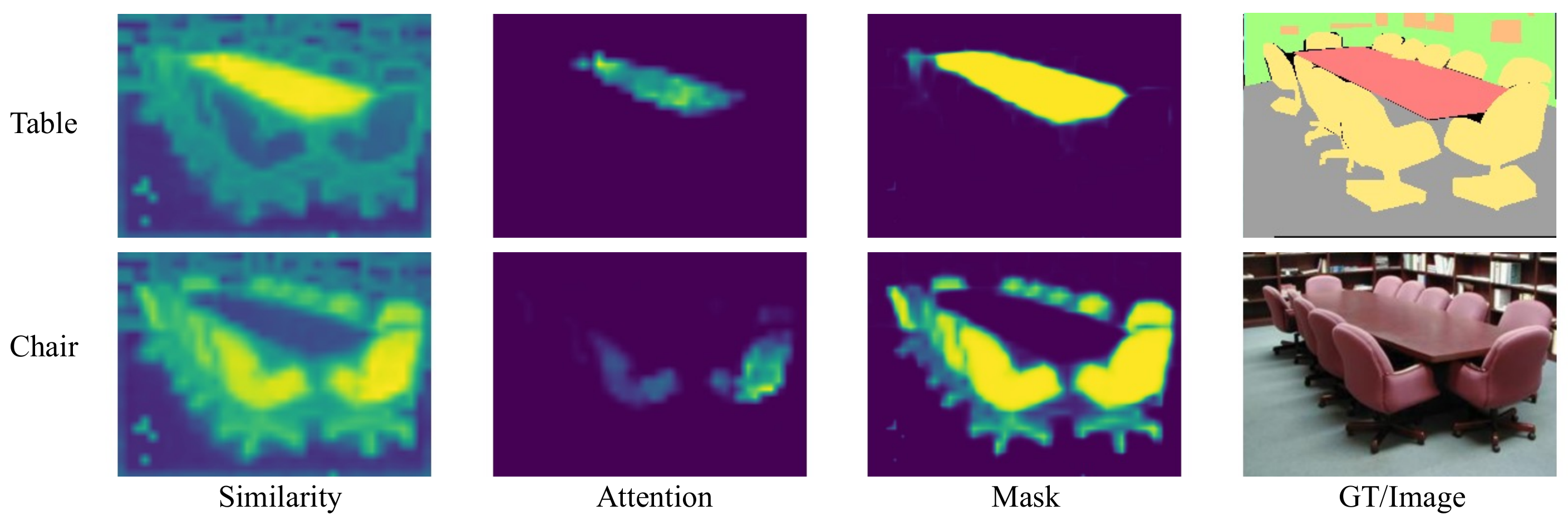}
    \caption{\textbf{The overall concept of 
    our Attention-to-Mask decoder.} 
    ATM learns the similarity map for each category by capturing the cross-attention between the class tokens and the spatial feature map (Left). ${\tt Sigmoid}$ is applied to produce category-specific masks, highlighting the area with high similarity to the corresponding class (Middle). ATM enhances the semantic representations by encouraging the feature to be similar to the target class token and dissimilar to other tokens. %
    }
    \label{fig:attn_vs_mask}
\end{figure*}

\section{Introduction}

Semantic segmentation is a pivotal computer vision task that aims to assign labels to every pixel on the image. 
Widely adopted state-of-the-art methods like Fully Convolutional Networks (FCN)~\cite{fcn} utilize deep convolutional neural networks (ConvNet) as encoders and incorporate segmentation decoders for dense predictions.
Prior works~\cite{wang2020deep, ocrnet, dv3} have aimed to enhance performance by augmenting contextual information or incorporating multi-scale information, leveraging the inherent multi-scale and \textit{hierarchical} attributes of the ConvNet architectures.

The advent of the Vision Transformer (ViT)~\cite{vit} has offered a paradigm shift, serving as a robust backbone for numerous computer vision tasks.ViT, distinct from ConvNet base models, retains a plain and \textit{non-hierarchical} architecture while preserving the resolution of the feature maps. To conveniently leverage existing segmentation decoders for dense prediction, such as U-Net~\cite{ronneberger2015u} or DeepLab~\cite{dv3}, recent Transformer-based approaches, including Swin Transformer~\cite{liu2021swin} and PVT~\cite{pvt}, have developed a \textit{hierarchical} ViT to extract hierarchical feature representations.

However, modifying the original ViT structures requires training the networks from scratch rather than using off-the-shelf plain ViT checkpoints due to the discrepancy between the hierarchical and plain architectures, such as spatial down-sampling~\cite{xu2022rethinking}.
Altering the plain ViT architecture compromises the use of rich representations from vision-language pre-training methods like CLIP~\cite{radford2021learning}, BEiT~\cite{beit}, BEiT-v2~\cite{beitv2}, MVP~\cite{wei2022mvp}, and COTS~\cite{lu2022cots}.

Hence, there is a clear advantage to developing effective decoders for the original ViT structures in order to leverage those powerful representations.
Previous works, such as UPerNet \cite{Upernet} and DPT \cite{DPT}, have primarily focused on hierarchical feature maps and neglected the distinctive characteristics of the plain Vision Transformer.
Consequently, these methods introduce computation-intensive operations while offering limited performance gains, as shown in Fig.~\ref{fig:intro_digraph_comparison}.

A recent trend in several works, such as SETR \cite{setr} or Segmenter \cite{strudel2021segmenter}, aims to develop decoders specifically tailored for the Plain ViT architecture.
However, these designs are often an extension of per-pixel classification techniques derived from traditional convolution-based decoders.
For example, SETR's decoder~\cite{setr} uses a sequence of convolutions and bilinear up-sampling to increase the ViT's extracted feature maps gradually.
 It then applies a naive MLP to the extracted features to perform pixel-wise classification, which isolates the neighboring contexts surrounding the pixel. 
 Current pixel-wise classification decoder designs overlook the importance of contextual learning when assigning labels to each pixel.

Another prevalent issue in deep networks, including Transformer, is `catastrophic forgetting'~\cite{french1999catastrophic, kirkpatrick2017overcoming}, where the model's performance on previously learned tasks deteriorates as it learns new ones~\cite{shao2022overcoming, wang2022learning, wang2022dualprompt,phan2022class}. This limitation poses significant challenges for the application of deep segmentation models in dynamic real-world environments.
Recently, the rapid development of the foundation model pre-trained on large-scale data has sparked interest among researchers in studying its transferability across various downstream tasks~\cite{ostapenko2022continual}.
These models are capable of extracting powerful and generalized representations, which has led to a growing interest in exploring their extensibility to new classes and tasks while retaining the previously learned knowledge representations~\cite{ramasesh2022effect, wu2022pretrained}.  %

Inspired by these challenges, this paper aims to develop plain Vision Transformer-based model for effective semantic segmentation without resorting to hierarchical backbone.
As self-supervision and multi-modality pre-training continue to evolve, we anticipate that the plain vision transformer will learn enhanced visual representations. Consequently, decoders for dense tasks are expected to adapt more flexibly and efficiently to these representations. %

In light of these research gaps, we propose \textbf{SegViTv2} — a novel, efficient segmentation network that features a plain Vision Transformer and exhibits robustness against forgetting. We introduce a novel Attention-to-Mask (ATM) module that operates as a lightweight component for the SegViT decoder. Leveraging the non-linearity of cross-attention learning, our proposed ATM employs learnable class tokens as queries to pinpoint spatial locations that exhibit high compatibility with each class. We advocate for regions affiliated with a particular class to possess substantial similarity values that correspond to the respective class token.

As depicted in~\cref{fig:attn_vs_mask}, the ATM generates a meaningful similarity map that accentuates regions with a strong affinity towards the `Table' and `Chair' categories. By simply implementing a {\tt Sigmoid} operation, we can transform these similarity maps into mask-level predictions. The computation of the mask scales linearly with the number of pixels, a negligible cost that can be integrated into any backbone to bolster segmentation accuracy. Building upon this efficient ATM module, we present a novel semantic segmentation paradigm that utilizes the cost-effective structure of plain ViT, referred to as SegViT. Within this paradigm, multiple ATM modules are deployed at various layers to extract segmentation masks at different scales. The final prediction is the summation of the outputs derived from these layers.

To alleviate the computational burdens of plain Vision Transformers (ViTs), we introduce the \emph{Shrunk} and \emph{Shrunk++} structures, which incorporate query-based downsampling (QD) and query-based upsampling (QU).  The proposed QD employs a 2x2 nearest neighbor downsampling technique to obtain a sparser token mesh, reducing the number of tokens involved in attention computations. In \textit{Shrunk++}, we extend QD to edge-aware query-based downsampling (EQD). EQD selectively preserves tokens situated at object edges, as they possess more discriminative information. Consequently, QU recovers the discarded tokens within the object's homogeneous body, reconstructing high-resolution features crucial for accurately dense prediction. Integrating the \emph{Shrunk++} structure with the ATM module as the decoder, our SegViTv2 achieves computational reductions of up to 50\% while maintaining competitive performance.

We further adapt our SegViTv2 framework for continual learning. Leveraging the robust, generalized representation of the foundational model, this paper investigates its adaptability to new classes and tasks, ensuring retention of prior knowledge. Recent techniques in continual semantic segmentation (CSS) aim to replay old data~\cite{maracani2021RECALLRC, cha2021ssul} or distill knowledge from the previous model to mitigate model divergence~\cite{cermelli2020ModelingTB, phan2022class, zhang2022representation}. These methods fine-tune parameters related to old tasks, which can disrupt the previously learned solutions and result in forgetting. In contrast, our proposed SegViT supports learning new classes without interfering with previously acquired knowledge. We strive to establish a forget-free SegViT framework, achieved by incorporating a new ATM module dedicated to new tasks while freezing all old parameters. Consequently, the proposed SegViT architecture has the potential to eliminate the issue of forgetting.

Our key contributions can be summarized as follows:

\begin{itemize}
    \item  We introduce the Attention-to-Mask (ATM) decoder module, a potent and efficient tool for semantic segmentation. For the first time, we exploit spatial information present in attention maps to generate mask predictions for each category, proposing a new paradigm for semantic segmentation.
    \item We present the \emph{Shrunk++} structure, applicable to any plain ViT backbone, which alleviates the intrinsically high computational expense of the \textit{non-hierarchical} ViT while maintaining competitive performance, as illustrated in Fig.~\ref{fig:intro_digraph_comparison}. We are the first work capitalizing on edge information to decrease and restore tokens for efficient computation. Our \emph{Shrunk++} version of \textbf{SegViTv2}, tested on the ADE20K dataset, achieves a mIoU of 55.7\%, with a computational cost of 308.8 GFLOPs, marking a reduction of approximately 50\% compared to the original SegViT (637.9 GFLOPs).
    \item We propose a new SegViT architecture capable of continual learning with nearly zero forgetting. To our knowledge, we are the first work seeking to completely freeze all parameters for old classes, thereby nearly obliterating the issue of catastrophic forgetting.
\end{itemize}

\section{Related Work}
\paragraph{Semantic Segmentation.}
Semantic segmentation aims to partition an image into regions with meaningful categories. Fully Convolutional Networks (FCNs) used to be the dominant approach to this task.
To enlarge the receptive field, several approaches \cite{pspnet, dv3} propose dilated convolutions or apply spatial pyramid pooling to capture contextual information at multiple scales.
 Most semantic segmentation methods aim to classify each pixel directly using a classification loss. This paradigm naturally partitions images into different classes. %

Various methods have achieved significant advancements by integrating Transformers into the semantic segmentation task. Early works~\cite{liu2021swin, dong2022cswin} directly adapt the transformer encoder, designed for classification, into semantic segmentation by fine-tuning it together with segmentation decoders such as UPerNet~\cite{Upernet}. Recent approaches~\cite{xie2021segformer, strudel2021segmenter, maskformer} have focused on designing the overall segmentation framework to achieve better adaptation. For instance, SETR~\cite{setr} views semantic segmentation as a sequence-to-sequence task and proposes a pure Transformer encoder combined with a standard convolution-based decoder. SegFormer~\cite{xie2021segformer} employs a hierarchical encoder design to extract features from fine-to-coarse levels and a lightweight decoder design for efficient prediction. However, the SegFormer decoder adopts the pyramid structure by fusing multi-scale features, which is specialized for hierarchical ViTs such as Swin Transformer~\cite{liu2021swin}. The above-mentioned methods aim to design either a naive convolution-based decoder or a pyramid-structure decoder for hierarchical base models.  Nonetheless, designing an effective decoder specialized for plain ViTs remains an open research question.

Recently, several segmentation methods propose a universal framework that unifies multiple tasks, including instance segmentation, semantic segmentation, and object detection.  For example, Mask DINO~\cite{li2022mask} extends DINO with a mask prediction branch, achieving promising results in the instance, panoptic, and semantic segmentation tasks. Mask2Former~\cite{cheng2021mask2former} enhances MaskFormer~\cite{maskformer} by introducing deformable multi-scale attention in the decoder and a masked cross-attention mechanism. OneFormer \cite{jain2022oneformer} represents a universal image segmentation framework with a multi-task train-once design, outperforming specialized models in various tasks.

Recent methods~\cite{maskformer, strudel2021segmenter, knet} propose decoupling the per-pixel classification into image partitioning and region classification. For image partitioning, they use learnable tokens as mask embeddings and associate them with the extracted feature map to generate object masks. For region classification, the learnable tokens are fed to a classifier to predict the class corresponding to each mask. %
This paradigm enables global segmentation and alleviates the burden on the decoder to perform per-pixel classification, resulting in state-of-the-art performance~\cite{maskformer}.  While previous works use generic tokens for mask generation, this work explicitly utilizes class-specific tokens to enhance the semantics of mask embeddings, thereby improving segmentation accuracy. %

\paragraph{Mask-oriented Segmentation.}
Compared to previous mask-oriented segmentation techniques such as MaskFormer \cite{cheng2021maskformer} and Mask2Former \cite{cheng2021mask2former}, our method presents several novel conceptual differences and advantages. Specifically, our approach is tailored to address semantic segmentation problems by assigning each class to a fixed token and generating the corresponding mask directly. In contrast, MaskFormer relies on Hungarian matching, with each learnable query corresponding to spatial information instead of category information. Our Attention-to-Mask (ATM) approach eliminates the need for positional embedding, as we utilize the attention map between the class token and the feature map. Our overarching goal is to adapt Plain Vision Transformers for dense prediction, as recent studies have demonstrated that self-supervised learning \cite{he2022masked,chen2022context,touvron2022deit,beitv2} and multimodal learning \cite{radford2021learning} are enhanced by hierarchical ViT structures. Our approach enhances the representation ability of class tokens by applying transformer blocks.

Previous CNN-based decoders, such as OCRNet \cite{yuan2019segmentation} and K-Net \cite{zhang2021k}, have demonstrated the effectiveness of the attention mechanism in modeling contextual information. For example, K-Net utilizes semantic kernels (one kernel for each class) and performs convolution operations to generate the semantic mask. In contrast, our proposed ATM module integrates cross-attention mechanisms, allowing for more effective contextual learning. While OCRNet~\cite{yuan2019segmentation} applies cross-attention from the class token to the feature map to enhance feature representations, it still employs a standard linear predictor in the decoder to produce the segmentation map. On the other hand, our proposed ATM module is specifically designed for generating segmentation outputs, paving the way for future research on effective decoders for plain ViT. Additionally, existing convolution-based attention networks such as OCRNet~\cite{yuan2019segmentation}, K-Net~\cite{zhang2021k}, and DANet~\cite{danet} adopt the traditional per-pixel classification framework for segmentation generation. In contrast, our proposed SegViT decouples segmentation into mask prediction and classification, which proves advantageous for establishing connections between the class proxy and language representations~\cite{zhou2022zegclip}, as well as facilitating continual learning.  %

\paragraph{Transformers for Vision.}
In the realm of image classification tasks, attention-based transformer models have emerged as powerful alternatives to standard convolution-based networks. The original ViT~\cite{vit} represents a plain, non-hierarchical architecture. However, there have been several advancements in the field of hierarchical transformers, such as PVT \cite{pvt}, Swin Transformer \cite{liu2021swin}, Twins \cite{chu2021twins}, SegFormer \cite{xie2021segformer}, and P2T \cite{p2t}.
These hierarchical transformer models inherit certain design elements from convolution-based networks, including hierarchical structures, pooling, and downsampling with convolutions. Consequently, they can be seamlessly employed as direct replacements for convolutional-based networks and can be coupled with existing decoder heads for tasks such as semantic segmentation.

\paragraph{Self-Supervised Vision Transformers.} Self-supervised learning has emerged as a powerful te\-ch\-ni\-q\-u\-e for pre\-tr\-ain\-ing visual models, eliminating the need for labeled data. One notable self-supervised method is MAE \cite{he2022masked} (Masked Autoencoder), which trains a vision transformer to reconstruct masked regions of input images. This approach results in a high generalization capacity. 

Another significant method is CLIP \cite{radford2021learning} (Contrastive Language-Image Pre-Training), which involves joint training of a vision transformer and a language model on a large corpus of text and images, leading to the creation of a comprehensive knowledge store. CAE \cite{chen2022context} aims to learn image representations that are invariant to context changes and effectively capture underlying semantic content. Furthermore, iBot \cite{zhou2021ibot} performs masked visual learning using an online tokenizer and self-distillation mechanism, facilitating semantic representation learning.

In our approach, we leverage attention to masks to optimize the extraction of dense hidden representations, thereby enhancing the segmentation capability of our model.

\begin{figure*}[h]
    \centering
    
    \includegraphics[width=0.908\textwidth]{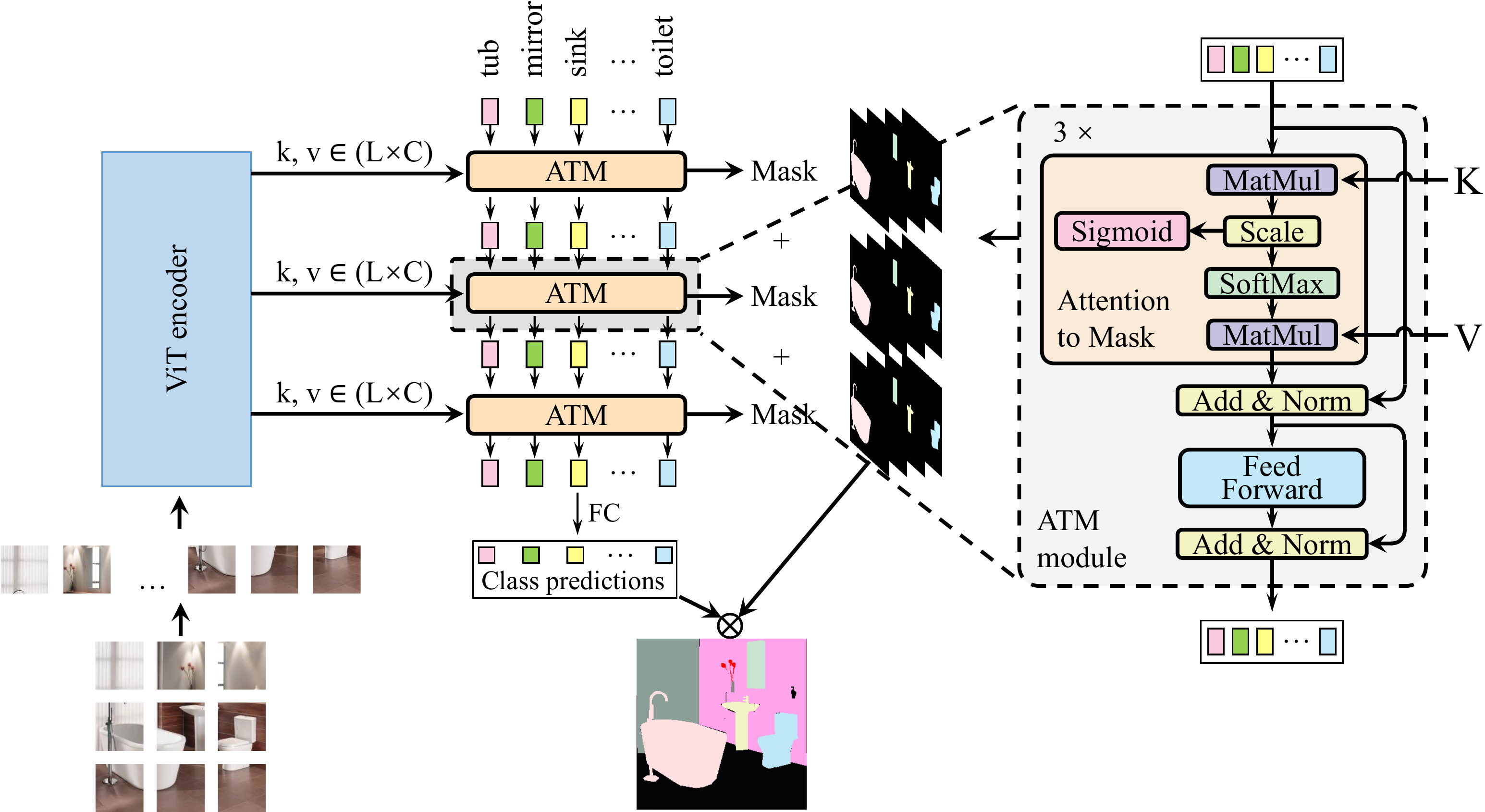}
    \caption{\textbf{The overall \seg\ structure with the \atm\ module.}
    The Attention-to-Mask (\atm) module  inherits the typical transformer decoder structure. It takes in  randomly initialized class embeddings as queries and the feature maps from the ViT backbone to generate keys and values. The outputs of the \atm\ module are used as the input queries for the next layer.
    The \atm\ module is carried out sequentially with inputs from different layers of the backbone as keys and values in a cascade manner. 
    A linear transform is then applied to the output of the 
    \atm\ module to produce the class predictions for each token. The mask for the corresponding class is transferred from the similarities between queries and keys in the \atm\ module. We have removed the self-attention mechanism in ATM decoder layers further improve the efficiency while maintaining the performance.
    }
    \label{fig:atm_decoder}
\end{figure*}

\paragraph{Plain-backbone decoders.}
For dense prediction tasks, such as semantic segmentation, the high-resolution feature maps produced by the backbone are vital for preserving spatial details. In typical hierarchical transformer models, techniques such as FPN \cite{fpn} or dilated backbone are employed to generate high-resolution feature maps by merging features from different levels. However, when it comes to a plain, non-hierarchical transformer backbone, the resolution remains the same across all layers.
SETR \cite{setr} proposed a straightforward approach to address segmentation tasks by treating transformer outputs from the base model in a sequence-to-sequence perspective. Segmenter \cite{strudel2021segmenter} combines class embeddings and transformer patch embeddings and applies several self-attention layers on the combined tokens to learn discriminative embeddings. In their approach, the class tokens are used as input to the ViT backbone, resulting in increased computational complexity.
In contrast, our SegViT introduces the class tokens as input to the ATM, the Attention-to-Mask module, thereby reducing computational costs while still benefiting from the integration of class tokens.

\paragraph{Continual Learning.} Continual learning (CL) aims to address the issue of forgetting, ensuring consistent performance on previously learned classes while adapting to new ones~\cite{chen2016LifelongML}. Most CL methods propose regularization techniques for convolution-based networks~\cite{li2018LearningWF,douillard2020podnet,kang2022class,peng2021hierarchical} or expand the network architectures to accommodate new tasks~\cite{yan2021dynamically}, thereby avoiding the need to store and replay old data.
In recent years, efforts have also emerged to prevent forgetting in Transformer models. Dytox~\cite{douillard2022dytox} dynamically learns new task tokens, which are then utilized to make the learned embeddings more relevant to the specific task. Lifelong ViT~\cite{wang2022continual} and contrastive ViT~\cite{wang2022online} introduce cross-attention mechanisms between tasks through external key vectors, and they slow down the changes to these keys to mitigate forgetting. Despite the use of complex mechanisms to prevent forgetting, these methods still require fine-tuning of the network for new classes, which can result in interference with previously learned knowledge.

In the field of semantic segmentation, recent research has been devoted to addressing the forgetting issue in continual learning. However, in addition to forgetting, continual semantic segmentation (CSS) also encounters the problem of "background shift." This refers to the situation where foreground object classes from previous tasks are mistakenly classified as background in the current task~\cite{cermelli2020ModelingTB}.
REMINDER~\cite{phan2022class} tackles forgetting in CSS by utilizing class similarity to identify the classes that are more likely to be forgotten. It then focuses on revising those specific classes to mitigate the forgetting problem. RCIL~\cite{zhang2022representation} introduces a two-branch convolutional network, with one branch frozen and the other trained to prevent forgetting. At the end of each learning step, the trainable branch is merged with the frozen branch, which can introduce model interference.
However, it is worth noting that existing CSS and CL techniques typically involve fine-tuning certain parts of the network dedicated to the old tasks. Unfortunately, this fine-tuning process can lead to forgetting as the model diverges from the previously learned solution. %

\section{Method}
 In this section, we first introduce the overall architecture of our proposed SegViT model for semantic segmentation. Then, we discuss the \emph{Shrunk} and \emph{Shrunk++} architectures designed to reduce the model's computational cost. Lastly, we explore the adaptation of our SegViT model for the context of continual semantic segmentation to minimize forgetting.

\subsection{Overall SegViT architecture}
SegViT comprises a ViT-based encoder responsible for feature extraction and a decoder used to learn the segmentation map. For the encoder, we designed the `Shrunk' structure to reduce the computational overhead associated with the plain ViT.
Regarding the decoder, we introduce a novel lightweight module named Attention-to-Mask (\atm). This module generates class-specific masks denoted as $M$ and class predictions denoted as $P$, which determine the presence of a particular class in the image. The mask outputs from a stack of ATM modules are combined and then multiplied by the class predictions to obtain the final segmentation output. Fig. \ref{fig:atm_decoder} illustrates the overall architecture of our proposed SegViT.

\subsubsection{Encoder}
Given an input image $I \in \mathbb{R}^{H \times W \times 3}$, the plain vision transformer backbone reshapes it into a sequence of tokens $\mathcal{F}_0 \in \mathbb{R}^{L \times C}$, where $L = \frac{HW}{P^2}$, $P$ is the patch size, and $C$ is the number of channels. 
To capture positional information, learnable position embeddings of the same size as $\mathcal{F}_0$ are added. Subsequently, the token sequence $\mathcal{F}_0$ is processed by $m$ transformer layers to produce the output.
The output tokens for each layer are defined as $[\mathcal{F}_1, \mathcal{F}_2, \dots, \mathcal{F}_m] \in \mathbb{R}^{L \times C}$.
For a plain vision transformer like ViT, the number of tokens are high and remains constant for each layer. Processing a substantial number of tokens for every layer results in elevated computational costs for plain ViT. We denote a plain ViT-based encoder as the 'Single' structure. To mitigate computational costs,  we introduce the \textit{Shrunk} and \textit{Shrunk++} structures, tailored to create a more computationally efficient ViT-based encoder. Further details regarding the \textit{Shrunk} structure can be found in Section~\ref{sec:shrunk}.

\subsubsection{Decoder}
\paragraph{Attention-to-Mask (\atm).}
Cross-attention can be described as the mapping between two sequences of tokens, denoted as $\{\mathbf{v_1}, \mathbf{v_2}\}$. In our case, we define two token sequences: $\mathcal{G} \in \mathbb{R}^{N \times C}$ with a length $N$ equal to the number of classes, and $\mathcal{F}_i \in \mathbb{R}^{L \times C}$.
To enable cross-attention, linear transformations are applied to each token sequence, resulting in the query (Q), key (K), and value (V) representations. This process is described by Equation~\eqref{eq:2}.
\begin{equation}
\centering
\begin{split}
    Q &  =  \phi_{q} (\mathcal{G}) \in \R^{N \times C},\\
    K &= \phi_{k} (\mathcal{F}_{i}) \in \R^{L \times C},\\ 
    V& = \phi_{v} (\mathcal{F}_{i}) \in \R^{L \times C}.
    \label{eq:2}
\end{split}
\end{equation}

The similarity map is calculated by computing the dot product between the query and key representations. Following the scaled dot-product attention mechanism, the similarity map and attention map are calculated as follows:
\begin{equation} 
\begin{split}
       S(Q, K) &= \frac{QK^T}{\sqrt{d_{k}}}\in \R^{N \times L}, \\
        Attention(\mathcal{G}, \mathcal{F}_{i}) & = 
        {\tt Softmax}\big(S(Q, K)\big)V \in \R^{N \times C}, 
\end{split}
\end{equation}
where $\sqrt{d_{k}}$ is a scaling factor with $d_{k}$ equals to the dimension of the keys. 

The shape of the similarity map $S(Q, K)$ is determined by the lengths of the two token sequences, $N$ and $L$. The attention mechanism updates $\mathcal{G}$ by performing a weighted sum of $V$, where the weights are derived from the similarity map after applying the softmax function along the $L$ dimension.

In dot-product attention, the softmax function is used to concentrate attention exclusively on the token with the highest similarity. However, we believe that tokens other than those with maximum similarity also carry meaningful information. Based on this intuition, we have designed a lightweight module that generates semantic predictions more directly.
To this end, we assign $\mathcal{G}$ as the class embeddings for the segmentation task, and $\mathcal{F}_i$ as the output of layer $i$ of the ViT backbone. A semantic mask is paired with each token in $\mathcal{G}$ to represent the semantic prediction for each class. The binary mask $M$ is defined as follows:
\begin{equation}
        Mask(\mathcal{G}, \mathcal{F}_{i}) = {\tt Sigmoid}(S(Q, K)) \in \R^{N \times L}.
\end{equation}
The masks have a shape of $N \times L$, which can be reshaped to $N \times \frac{H}{P} \times \frac{W}{P}$ and bilinearly upsampled to the original image size $N \times H \times W$. As depicted in the right section of Fig. \ref{fig:atm_decoder}, the \atm\ mechanism produces masks as an intermediate output during cross-attention. 

The final output tokens $Z \in \mathbb{R}^{L \times C}$ from the ATM module are utilized for classification. A fully connected layer (FC) parameterized by $W \in \mathbb{R}^{C \times 2}$ followed by the Softmax function is used to predict whether the object class is present in the image or not. The class predictions $\mathcal{P} \in \mathbb{R}^{N \times 2}$ are formally defined as:

\begin{equation}
\centering
\label{eq:cls}
\begin{aligned} 
    \mathcal{P} = \text{Softmax}(WZ).
    \end{aligned}
\end{equation}
Here, $P_{c,1}$ indicates the likelihood of class $c$ appearing in the image. For simplicity, we refer to $P_c$ as the probability score for class $c$. 

The output segmentation map for class $O_s \in \mathbb{R}^{H \times W}$ is obtained by element-wise multiplication of the reshaped class-specific mask $M_c$ and its corresponding prediction score $P_c$: $O_c = P_c \odot M_c$. During inference, the label is assigned to each pixel $i$ by selecting the class with the highest score using $\text{argmax}_{c} O_{i,c}$.

Indeed, plain base models like ViT do not inherently possess multiple stages with features of different scales. Consequently, structures such as Feature Pyramid Networks (FPN) that merge features from multiple scales are not applicable to them.

Nevertheless, features from layers other than the last one in ViT contain valuable low-level semantic information, which can contribute to improving performance. In SegViT, we have developed a structure that leverages feature maps from different layers of ViT to enrich the feature representations. This allows us to incorporate and benefit from the rich low-level semantic information present in those feature maps.

SegViT is trained via the classification loss and the binary mask loss. The classification loss ($\mathcal{L}_{\text{cls}}$) minimizes cross-entropy between the class prediction and the actual target. The mask loss ($\mathcal{L}_{\text{mask}}$) consists of a focal loss \cite{lin2017focal} and a dice loss \cite{diceloss} for optimizing the segmentation accuracy and addressing sample imbalance issues in mask prediction. The dice loss and focal loss respectively minimize the dice and focal scores between the predicted masks and the ground-truth segmentation. The final loss is the combination of each loss, formally defined as:

\begin{equation}
\label{eq:loss}
\mathcal{L}=\mathcal{L}_{\text{cls}} +\lambda_{\text{focal}}\mathcal{L}_{\text{focal}} + \lambda_{\text{dice}}\mathcal{L}_{\text{dice}}
\end{equation}
where $\lambda_{\text{focal}}$ and $\lambda_{\text{dice}}$ are hyperparameters that control the strength of each loss function. Previous mask transformer methods such as MaskFormer \cite{maskformer} and DETR \cite{detr} have adopted the binary mask loss and fine-tuned their hyperparameters through empirical experiments. Hence, for consistency, we directly use the same values as MaskFormer and DETR for the loss hyperparameters: $\lambda_{\text{focal}}=20.0$ and $\lambda_{\text{dice}}=1.0$.

\subsection{ \emph{Shrunk} Structure for Efficient Plain ViT Encoder}
\label{sec:shrunk}
Recent efforts, such as DynamicViT~\cite{rao2021dynamicvit}, TokenLearner~\cite{ryoo2021tokenlearner}, and SPViT~\cite{kong2022spvit}, propose token pruning techniques to accelerate vision transformers. However, most of these approaches are specifically designed for image classification tasks and, as a result, discard valuable information. However, when adapting these techniques to semantic segmentation tasks, they may fail to preserve high-resolution features that are necessary for accurate dense prediction tasks.

\begin{figure}
    \centering
\includegraphics[width=0.46\textwidth]{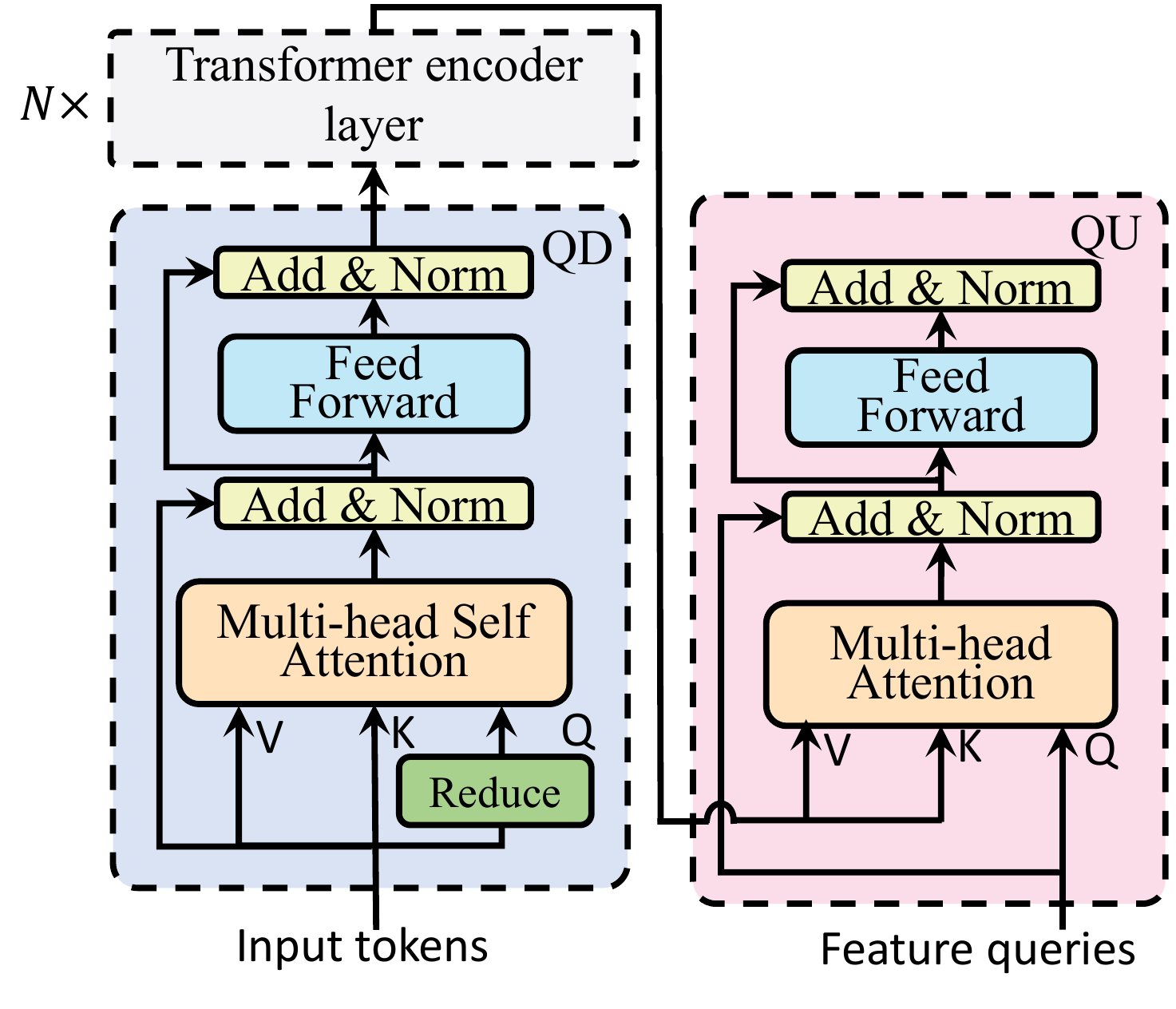}
    \caption{Architecture of the proposed query-downsapling (QD) layer (\textbf{\textcolor{blue}{blue}} block) and the query-upsampling (QU) layer (\textbf{\textcolor{pink}{pink}} block). The QD layer uses an efficient down-sampling technique (\textbf{\textcolor[HTML]{A9D18E}{green}} block) and removes less informative input tokens used for the query.  The QU layer takes a set of trainable query tokens and learns to recover the discarded tokens using multi-head attention. %
    }
    \label{fig:qu_qd}
\end{figure}
In this paper, we introduce the \emph{Shrunk} structure. This method employs query-based down-sampling (QD) to prune the input token sequence $\mathcal{F}_i$ and uses query up-sampling (QU) to retrieve the discarded tokens, ensuring preservation of fine-detail features vital for semantic segmentation.
The overall architecture of QD and QU is illustrated in Fig.~\ref{fig:qu_qd}. %

For QD, we have re-designed the Transformer encoder block~\cite{2017attention} and incorporated efficient down-sampling operations to specifically reduce the number of query tokens. In a Transformer encoder layer, the computational cost is directly influenced by the number of query tokens, and the output size is determined by the query token size. To mitigate the computational burden while maintaining information integrity, a viable strategy is to selectively reduce the number of query tokens while preserving the key and value tokens. This approach allows for an effective reduction in the output size of the current layer, leading to reduced computational costs for subsequent layers.

For QU, we perform up-sampling using a token sequence — either predefined or inherited — that has a higher resolution than the query tokens.
The key and value tokens are taken from the token sequence obtained from the backbone, which typically has a lower resolution. The output size is dictated by the query tokens with higher resolution. Through the cross-attention mechanism, information from the key and value tokens is integrated into the output. This process facilitates a non-linear merging of information and demonstrates an upsampling behavior, effectively increasing the resolution of the output.

\begin{figure*}[t]
    \centering
    \includegraphics[width=0.908\textwidth]{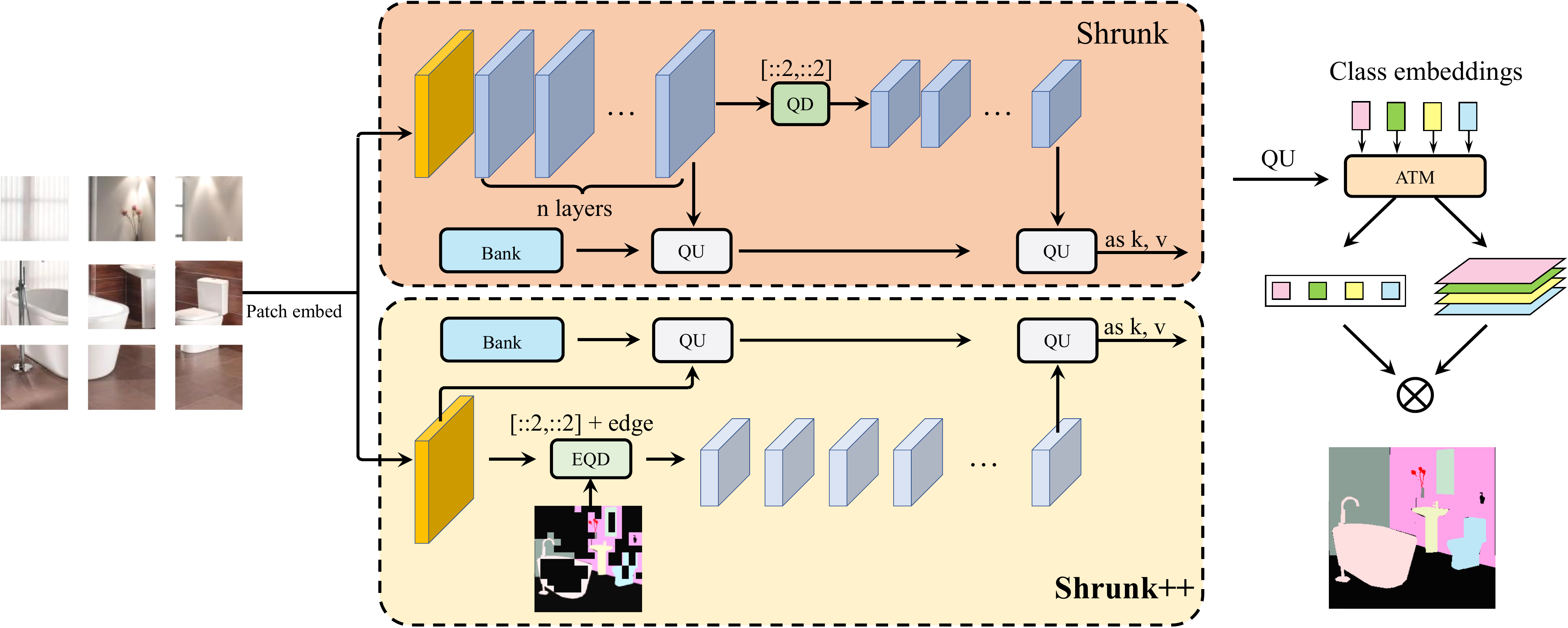}
    \caption{
    \textbf{Illustrations of the \emph{Shrunk} and \emph{Shrunk++}.
    } In the diagram, the \textcolor{blue}{blue} and \textcolor{orange}{orange} boxes respectively refer to the transformer encoder block and the patch embedding block. In SegVit~\cite{zhang2022segvit}, the proposed \emph{Shrunk} structure employs query downsampling (QD) on the middle-level features to preserve the information.  In the new \emph{Shrunk++} architecture, we introduce the Edged Query Downsampling (EQD) technique which consolidates every four adjacent tokens into one token and additionally includes the tokens that contain edges. 
    This enhancement enables downsampling operations to take place before the first layer without significant performance degradation, offering computational savings for the initial layers of the \emph{Shrunk} model. The edge information is extracted using a lightweight parallel edge detection head.}
    \label{fig:struct}
\end{figure*}

As illustrated in Fig. \ref{fig:struct}, our proposed \emph{Shrunk} structure incorporates the QD and QU modules. Specifically, we integrate a QD operation at the middle depth of the ViT backbone, precisely at the $8^\text{th}$ layer of a $24$-layer backbone. The QD operation downsamples the query tokens using a $2\times 2$ nearest neighbor downsampling operation, resulting in a feature map size reduction to $\nicefrac{1}{32}$. However, such downsampling can potentially cause information loss and performance degradation. To mitigate this issue, prior to applying the QD operation, we employ a QU operation to the feature map. This involves initializing a set of query tokens with a resolution of $\nicefrac{1}{16}$ to store the information. Subsequently, as the downsampled feature map progresses through the remaining backbone layers, it is merged and upsampled using another QU operation alongside the previously stored $\nicefrac{1}{16}$ high-resolution feature map. This iterative process ultimately generates a $\nicefrac{1}{16}$ high-resolution feature map enriched with semantic information processed by the backbone.

Despite the effectiveness of the proposed \emph{Shrunk} approach in maintaining performance, it requires the integration of the QD operation within the intermediate layers of the backbone. This necessity arises due to the fact that shallow layers primarily capture low-level features, and applying downsampling to these layers would result in significant information loss. Consequently, these low-level layers continue to be computed at a higher resolution, limiting the potential reduction in computational cost.

To address this limitation and further optimize the backbone, we introduce SegViTv2 using a novel architecture called \emph{Shrunk++}. In this architecture, we incorporate an edge detection module in the QD section and introduce an Edged Query Downsampling (EQD) technique to update the QD process. In addition to the $2\times2$ nearest downsampling operation that eliminates every 4 consecutive tokens, our approach aims to retain tokens that contain multiple categories, specifically tokens that contain an edge. By preserving the $2\times2$ sparse tokens, we retain important semantic information, while also preserving the edge tokens to retain detailed spatial information. By retaining both types of information, we minimize the loss of valuable information and overcome the limitations associated with low-level layers.
To extract edges, we add a separate branch using a lightweight multilayer perceptron (MLP) termed as the edge detection head that learns to detect edges from the input image. 
The edge detection head operates as an auxiliary branch, trained simultaneously with the main ATM decoder. This head processes the input image, which has the same dimensions as the backbone. Let the input image have $C$ channels, aligned with the backbone. The Multi-Layer Perceptron (MLP) in this head consists of three layers, with dimensions $C$, $C/2$, and $2$, respectively. Let $I$ represent the input image, and the output of the MLP can be defined as $E = \text{MLP}(I; W_1, W_2, W_3)$, where $W_1, W_2, W_3$ are the weights for the three layers. The output $E$ is then passed through a softmax activation function, resulting in $S = \text{Softmax}(E)$. To determine the confidence level of a token belonging to an edge, we apply a threshold $\tau$. In our implementation, we set $\tau$ to 0.7.
To obtain the ground-truth (GT) edge, we perform post-processing on the GT segmentation map $Y$. Since the input has been tokenized with a patch size of $P$, we tokenize the GT and reshape it into a sequence of tokens denoted as $Y \in R^{(HW/P^2) \times P \times P}$, where the last two dimensions correspond to the patch dimensions. We consider a patch to contain an edge if there exists any edge pixel within the patch. We define the edge mask $Mask_i$ as follows:

\begin{equation}
Mask_{i} =
\begin{cases}
1 & \text{if } \sum_{j,k} Y_{i,j,k} > 0, \\
0 & \text{otherwise}.
\end{cases}
\end{equation}

For each element $s_i$ in $S$, we create a binary edge mask $M_i$: $M_i = 1, \text{if } s_i \ge \tau$.
The cross-entropy loss is computed between the generated edge mask $M_i$ and the ground-truth edge mask $Y_i$:
$\mathcal{L}_{\text{edge}} = - \sum{i} Y_i \log(M_i) + (1 - Y_i) \log(1 - M_i)$.
By incorporating the Edge Detection head as an auxiliary branch, the Shrunk++ architecture effectively retains detailed spatial contexts throughout the query downsampling process, forming an Edge Query Downsampling (EQD) structure. This EQD structure effectively captures and retains edge information during sparse downsampling, significantly reducing computational overhead while maintaining performance.
The integration of EQD enables the Shrunk++ architecture to strike a remarkable balance between computational efficiency and maintaining high-performance levels.

\subsection{Exploration on Continual Semantic Segmentation}

Continual semantic segmentation aims to train a segmentation model in $T$ steps without forgetting. At step $t$, we are given a dataset $\cD^t$ which comprises a set of pairs $(X^t, Y^t)$, 
where $X^t$ is an image of size $H\times W$ and  $Y^t$ is the ground-truth segmentation map.
Here, $Y^t$ only consists of labels in current classes $\cC^{t}$, while all other classes (i.e., old classes $\cC^{1:t-1}$ or future classes $\cC^{t+1:T}$) are assigned to the background. In continual learning, the model at step $t$ should be able to predict all classes $\cC^{1:t}$ in history.

\begin{figure*}[h]
    \centering
\includegraphics[width=0.80048\textwidth]{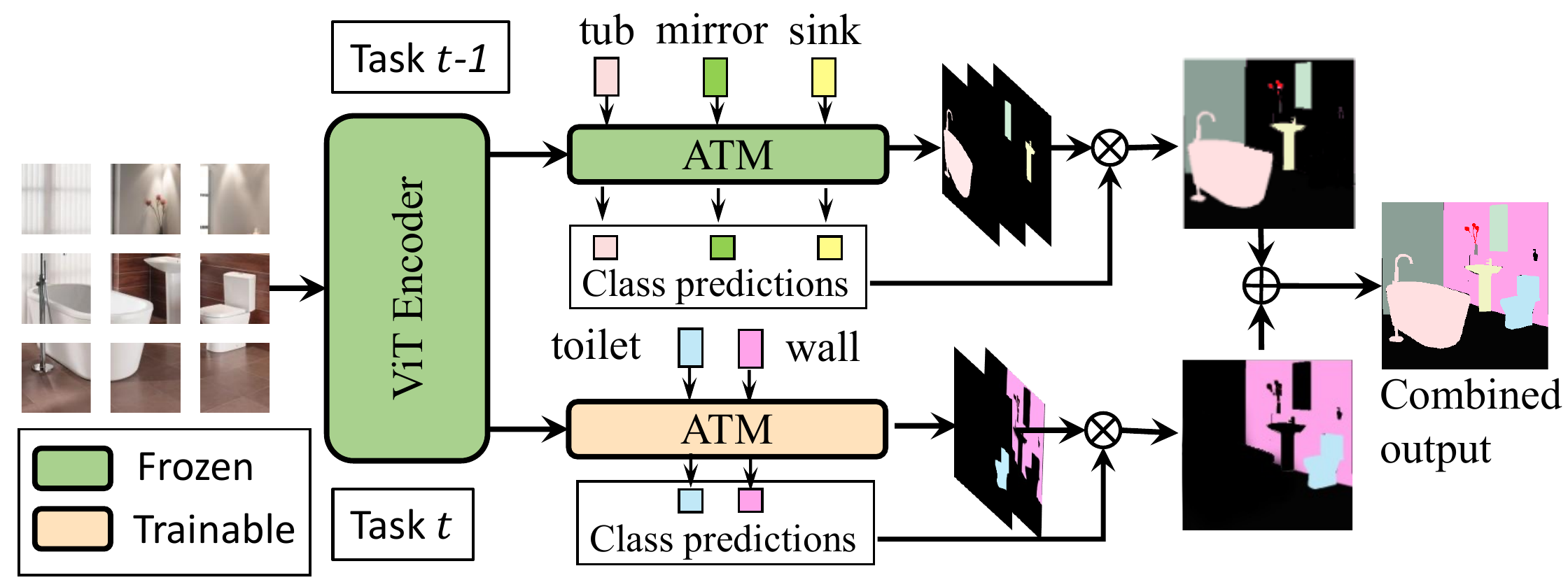}
    \caption{Overview of SegViT adapted for continual semantic segmentation. When learning a new task $t$, we grow and train a separate ATM and fully-connected layer to produce mask and class prediction. All the parameters dedicated to the old task $t-1$, including ATM, FC layers, and the ViT encoder, are frozen. This prevents interfering with the old knowledge, which guarantees no forgetting.}
    \label{fig:cl}
\end{figure*}

\textbf{SegViT for Continual Learning.} Existing continual semantic segmentation methods~\cite{zhang2022representation,phan2022class} propose regularization algorithms to preserve the past knowledge of a specific architecture, DeepLabV3. These methods focus on continual semantic segmentation for DeepLabV3 with a ResNet backbone, which has a less robust visual representation for distinguishing between different categories. Consequently, these methods require fine-tuning model parameters to learn new classes while attempting to retain knowledge of old classes. Unfortunately, adapting the old parameters dedicated to the previous task inevitably interferes with past knowledge, leading to catastrophic forgetting.
In contrast, our proposed SegViT decouples class prediction from mask segmentation, making it inherently suitable for a continual learning setting. By leveraging the powerful representation capability of the plain vision transformer, we can learn new classes by solely fine-tuning the class proxy (i.e., the class token) while keeping the old parameters frozen. This approach eliminates the need for fine-tuning old parameters when learning new tasks, effectively addressing the issue of catastrophic forgetting.

During training on the current task $t$, we add a new sequence of learnable tokens $\mathcal{G}^{t} \in \R^{|\cC^{t}| \times C}$, where $|\cC^{t}|$ is the number of classes in the current task.
To learn new classes, we grow and train new ATM modules and a fully-connected layer for mask prediction and mask classification. For simplicity, we ignore the parallel structure of ATM modules. A single ATM module refers to multiple ATM modules. Let $A^{t}$ and $W^{t}$ denote the ATM module and the weights of the fully connected  (FC) layer for task $t$. All parameters for prior tasks, including the ViT encoder, the ATM module, and the FC layer, are completely frozen. Fig.~\ref{fig:cl} illustrates the overview of our SegViT architecture adapted for continual semantic segmentation.

Given the encoder extracted features $\mathcal{F}_{T}$ and the class tokens $\mathcal{G}^{t}$, the ATM produces the mask predictions $M^{t}$ and the output tokens $Z^{t}$ corresponding to the mask:
\begin{equation}
      M^{t}, Z^{t} = \text{ATM}(\mathcal{G}^{t}, \mathcal{F}^{T}).
\end{equation}
Based on Eq.~\ref{eq:cls}, the class prediction $\mathcal{P}$ is obtained by applying FC on the class token $Z^t$. 

The prediction score $S^t_c$ for each class $c$ is multiplied by the corresponding mask $M^t_c$ to get the segmentation map $O^t_c$ for class $c$:
\begin{equation}
O^t_c = S^t_c \odot M^t_c,
\end{equation}
where $\odot$ denotes the element-wise multiplication.
The segmentation $\hat{O}^t$ is obtained by taking the class $c$ having the highest score in every pixel, defined as
\begin{equation}
\label{eq:seg}
    \hat{O}^t = \underset{c \in \cC^t}{\text{argmax }} O^t_{i,c}
\end{equation}
Based on the ground truth $Y^t$ for task $t$, SegViT is trained using the loss function defined in Eq.~\ref{eq:loss}. To produce the final segmentation across all tasks, we concatenate the individual outputs $O^t$ from each task. %

\section{Experiments}
\label{sec:exp}
\subsection{Datasets}
\textbf{ADE20K \cite{ade20k}}  
is a challenging scene parsing dataset which contains $20,210$ images  as the training set and $2,000$ images as the validation set with 150 semantic classes. 

\noindent\textbf{COCO-Stuff-10K \cite{cocostuff}}
is a scene parsing benchmark with $9,000$ training images and $1,000$ test images. Even though the dataset contains $182$ categories, not all categories exist in the test split. We follow the implementation of mmsegmentation \cite{mmseg} with $171$ categories to conduct the experiments.

\noindent\textbf{PASCAL-Context \cite{pascal_context}}
is a dataset with $4,996$ images in the training set and $5,104$ images in the validation set. There are $60$ semantic classes in total, including a class representing `background'. 

\begin{table*}[!ht]
    \centering
        \caption{Experiment results on the ADE20K \texttt{val.}\  split.  `ms' means that mIoU is calculated using multi-scale inference. `$\dagger$' means the models use the backbone weights pre-trained by AugReg \cite{augreg}. `*' represents the model reproduced under the same settings as the official repo. The GFLOPs are measured at single-scale inference with the given crop size.
        We report inference speed for our SegViT and reproduce previous methods in terms of Frame Per Second (FPS) on a single A100 device. 
    }
    \vspace{0.5em}
    \begin{tabular}{rlccccc}
        \toprule
        Method & Backbone & Crop Size & GFLOPs & mIoU (ss) & mIoU (ms) & Inf time (fps) \\
        \midrule
        UPerNet  \cite{Upernet} & ViT-Base & $512 \times 512$ & 443.9 & 46.6 & 47.5 & 16.07 \\
        DPT* \cite{DPT} &ViT-Base& $512 \times 512$ & 219.8 &47.2 & 47.9 & 23.63 \\
        SETR-MLA* \cite{setr} &ViT-Base& $512 \times 512$ & 113.5 &48.2& 49.3 & - \\
        Segmenter* \cite{strudel2021segmenter}&ViT-Base & $512 \times 512$& 129.6 &49.0 & 50.0 & 20.46 \\
        StructToken \cite{lin2022structtoken}  & ViT-Base  & $512 \times 512$& 171.5 & 50.9 & 51.8 & 14.22\\
        MaskFormer \cite{cheng2021maskformer} & Swin-B(21K) & $640 \times 640$ & 198.3 & 52.7 & 53.9 & - \\
        Mask2Former \cite{cheng2021mask2former} & Swin-B(21K) & $640 \times 640$ & 223.4 & \textit{53.9} & \textbf{55.1} & 12.43 \\
        \midrule
        \seg\ (Ours) &ViT-Base & $512 \times 512$& 120.9 & 51.3 & 53.0 & 31.52 \\
        \seg\ (\emph{Shrunk++}, Ours) &BEiTv2-Base & $512 \times 512$& \textbf{74.4} & 52.9 & 53.3 & 25.03 \\
        \seg\ (Ours) &BEiTv2-Base & $512 \times 512$& 120.9 & \textbf{54.0} & \textit{54.9} & 23.59 \\
        \midrule
        \midrule
        
        DPT* \cite{DPT} & ViT-Large\textsuperscript{\textdagger} & $640 \times 640$ & 800.0 & 49.2& 49.5 & 9.38 \\
		UPerNet
         \cite{Upernet} & ViT-Large\textsuperscript{\textdagger}& $640 \times 640$ & 1993.9  & 48.6& 50.0 & 3.88\\
        SETR-MLA \cite{setr} & ViT-Large & $512 \times 512$ & 368.6 & 48.6&50.3 & 5.17\\
        MCIBI \cite{MCIBI}   & ViT-Large & $512 \times 512$ & \textgreater400 & - &50.8 & - \\
        Segmenter \cite{strudel2021segmenter} & ViT-Large\textsuperscript{\textdagger}& $640 \times 640$ & 671.8 & 51.8 & 53.6 & 4.73\\
        StructToken \cite{lin2022structtoken}  & ViT-Large\textsuperscript{\textdagger} & $640 \times 640$ & 774.6 & 52.8 & 54.2 & 4.1 \\
        KNet+UPerNet \cite{knet} & Swin-L(21K) & $640 \times 640$ & 659.3 & 52.2 & 53.3 & 11.28 \\
        MaskFormer \cite{cheng2021maskformer} & Swin-L(21K) & $640 \times 640$ & 378.1 & 54.1 & 55.6 & 10.21\\
        Mask2Former \cite{cheng2021mask2former} & Swin-L(21K) & $640 \times 640$ & 402.7 & 56.1 & 57.3 & 8.81\\
        \midrule
       \seg\ (ours) & ViT-Large\textsuperscript{\textdagger}& $640 \times 640$ & 637.9 &54.6 & 55.2 & 9.37\\
        \seg (\emph{Shrunk} , ours) & ViT-Large\textsuperscript{\textdagger}& $640 \times 640$ & 373.5 & 53.9 & 55.1 & 10.18 \\
        \seg (\emph{Shrunk++}, ours) & ViT-Large\textsuperscript{\textdagger}& $640 \times 640$ & 209.1 & 53.0 & 54.9 & 10.26 \\
        \seg\ (\emph{Shrunk++}, ours) & BEiTv2-Large& $512 \times 512$ & 210.3 &55.1 & 56.1 & 9.82\\
        \seg\ (ours) & BEiTv2-Large& $512 \times 512$ & 374.0 & \textit {56.5} & \textit{58.0} & 9.39 \\
        \seg\ (\emph{Shrunk++}, ours) & BEiTv2-Large& $640 \times 640$ & 308.8 &55.7 & 57.0 & 9.38 \\
        \seg\ (ours) & BEiTv2-Large& $640 \times 640$ & 637.9 &\textbf{58.0} & \textbf{58.2} & 6.25\\

        \bottomrule
    \end{tabular}
    \label{tab:ade20k}
\end{table*}

\subsection{Implementation details}

\textbf{Transformer backbone.} 
We employ the naive ViT~\cite{vit} as the backbone for our method. 
For our ablation studies, we primarily utilize the `Base' variation, while also presenting results based on the `Large' variant. Notably, variations in performance can arise due to different pre-trained weights, as indicated by Segmenter~\cite{strudel2021segmenter}. To ensure equitable comparisons, we adopt the pre-trained weights provided by Augreg~\cite{augreg}, aligning with practices employed in Strudel~\cite{strudel2021segmenter} and StructToken~\cite{lin2022structtoken}. These weights stem from training on ImageNet-21k with strong data augmentation and regularization techniques~\cite{augreg}.
To explore the maximum capacity and assess the upper bound of our method, we also conduct experiments using stronger base models such as DEiT v3~\cite{touvron2022deit} and BEiT v2~\cite{beitv2}.

\noindent\textbf{Training settings.}
We use MMSegmentation \cite{mmseg} and follow the commonly used training settings. 
During training, we apply sequential data augmentation techniques, including random horizontal flipping, random resizing within a ratio of $0.5$ to $2.0$, and random cropping. For most settings, the cropping dimensions are set to $512\times 512$, except for PASCAL-Context where we use $480 \times 480$, and for ViT-large backbone on ADE20K where we use $640\times 640$.
The batch size is set to $16$ for all datasets with a total iteration of $160k$, $80k$, and $80k$ for ADE20k, COCO-Stuff-10k, and PASCAL-Context respectively.

\begin{table*}[!h]
\centering
\caption{Experiment results on the COCO-Stuff-10K \texttt{test.}\  split. Following 
  published methods, 
  we report the results with multi-scale inference (denoted by `ms'). %
  The GFLOPs is measured at single scale inference with a crop size of $512 \times 512$. 
}
\vspace{0.5em}
\begin{tabular}{rlcc}
\toprule
Method                     & Backbone & GFLOPs   & mIoU (ms) \\
\midrule
DANet        \cite{danet}              & Dilated-ResNet-101  & 289.3 &39.7      \\
MaskFormer    \cite{maskformer}             & ResNet-101-fpn  & 81.7 &  39.8      \\
EMANet      \cite{emanet}               & Dilated-ResNet-101 & 247.4  & 39.9      \\
SpyGR        \cite{spygr}              & ResNet-101-fpn &  \textgreater80 & 39.9      \\
OCRNet       \cite{ocrnet}              & HRNetV2-W48 &167.9 & 40.5      \\
GINet       \cite{wu2020ginet}               & JPU-ResNet-101 &  \textgreater200 & 40.6      \\
RecoNet     \cite{reconet}               & Dilated-ResNet-101  &  \textgreater200 & 41.5      \\
ISNet     \cite{jin2021isnet}                 & Dilated-ResNeSt-101 & 228.3  &42.1      \\

MCIBI   \cite{MCIBI}                   & ViT-Large  &  \textgreater380     & 44.9      \\
StructToken \cite{lin2022structtoken}               & ViT-Large  &  \textgreater400     & 49.1      \\
SenFormer \cite{bousselham2021efficient}               & Swin-Large &  \textgreater400     & 50.1      \\
\midrule
\seg\ (\emph{Shrunk}, ours) & ViT-Large   & 224.8    &  49.40      \\
\seg\ (ours) & ViT-Large   & 383.9   & 50.30 \\
\seg\ (\emph{Shrunk++}, ours) & BEiTv2-Large  & \textbf{213.3}    &  50.54 \\
\seg\ (ours) & BEiTv2-Large   & 388.2  & \textbf{53.46} \\
\bottomrule
\end{tabular}
\label{tab:cocostuff}
\end{table*}

\begin{table*}[!htbp]
\centering
\caption{Experimental results on the PASCAL-Context \texttt{val.}\  split. Following 
  published methods, 
  we report the results with multi-scale inference (denoted by `ms'). 
   mIoU$_{59}$: mIoU averaged over $59$ classes (without background). mIoU$_{60}$: mIoU averaged over $60$ classes ($59$ classes plus background). Both metrics were used in the literature, and we report for the $60$ classes. 
  The GFLOPs are measured at single scale inference with a crop size of $480 \times 480$.  
  }
  \vspace{0.5em}
\label{tab:pascal}
\begin{tabular}{rlc cc}
\toprule
Method     & Backbone & GFLOPs  &  mIoU$_{59}$  (ms) & mIoU$_{60}$  (ms)  \\
\midrule
    RefineNet   \cite{ lin2017refinenet}& ResNet-152 & -& - & 47.3\\
    UNet++  \cite{zhou2018unet++}  & ResNet-101 & -& 47.7 &-\\
    PSPNet   \cite{pspnet}&Dilated-ResNet-101 &  157.0 & 47.8 &- \\
    Ding \textit{et al.}  \cite{ding2018context} & ResNet-101 & -&51.6&-\\
    EncNet  \cite{Encnet}   &Dilated-ResNet-101& 192.1 &52.6 &- \\
    HRNet \cite{hrnet}   & HRNetV2-W48& 82.7 & 54.0 & 48.3 \\
    NRD  \cite{nrd} & ResNet-101 & 42.9 & 54.1 & 49.0 \\
    GFFNet \cite{li2020gated} & {\color{black}{Dilated-ResNet-101}} & {\color{black}{-}}& {\color{black}{54.3}} &{\color{black}{-}}\\

   EfficientFCN \cite{liu2020efficientfcn}  & {\color{black}{ResNet-101}} & {\color{black}{52.8}}& {\color{black}{55.3}} &{\color{black}{-}}\\
    
    {\color{black}{OCRNet  \cite{ocrnet}}} & {\color{black}{HRNetV2-W48}} & {\color{black}{143.9}}& {\color{black}{56.2}} &{\color{black}{-}}\\
    
SETR-MLA  \cite{setr}  & ViT-Large  & 318.5 & -     & 55.8      \\

Segmenter \cite{strudel2021segmenter}  & ViT-Large    & 346.2  & - &59.0     \\
SenFormer \cite{bousselham2021efficient}  & Swin-Large    & -  & 64.0 &   -   \\

\midrule
\seg\ (\emph{Shurnk}, ours) & ViT-Large  & 186.9 & 62.3 & 57.40\\
\seg\ (ours) & ViT-Large & 321.6 & 65.3 & 59.30 \\
\seg\ (\emph{Shurnk++}, ours) & BEiTv2-Large & \textbf{179.3} & 64.91 & 59.92 \\
\seg\ (ours) & BEiTv2-Large & 329.7 & \textbf{67.14} & \textbf{61.63} \\
\bottomrule
\end{tabular}
\end{table*}

\noindent\textbf{Evaluation metric.} We use the mean Intersection over Union (mIoU) as the metric to evaluate the performance. `ss' means single-scale testing and `ms' test time augmentation with multi-scaled $(0.5, 0.75, 1.0, 1.25, 1.5, 1.75)$ inputs. All reported mIoU scores are in a percentage format. All reported computational costs in GFLOPs are measured using the fvcore\footnote{\url{https://github.com/facebookresearch/fvcore}} library.

\begin{figure*}[!ht]
    \centering
    \includegraphics[width=.9\linewidth]{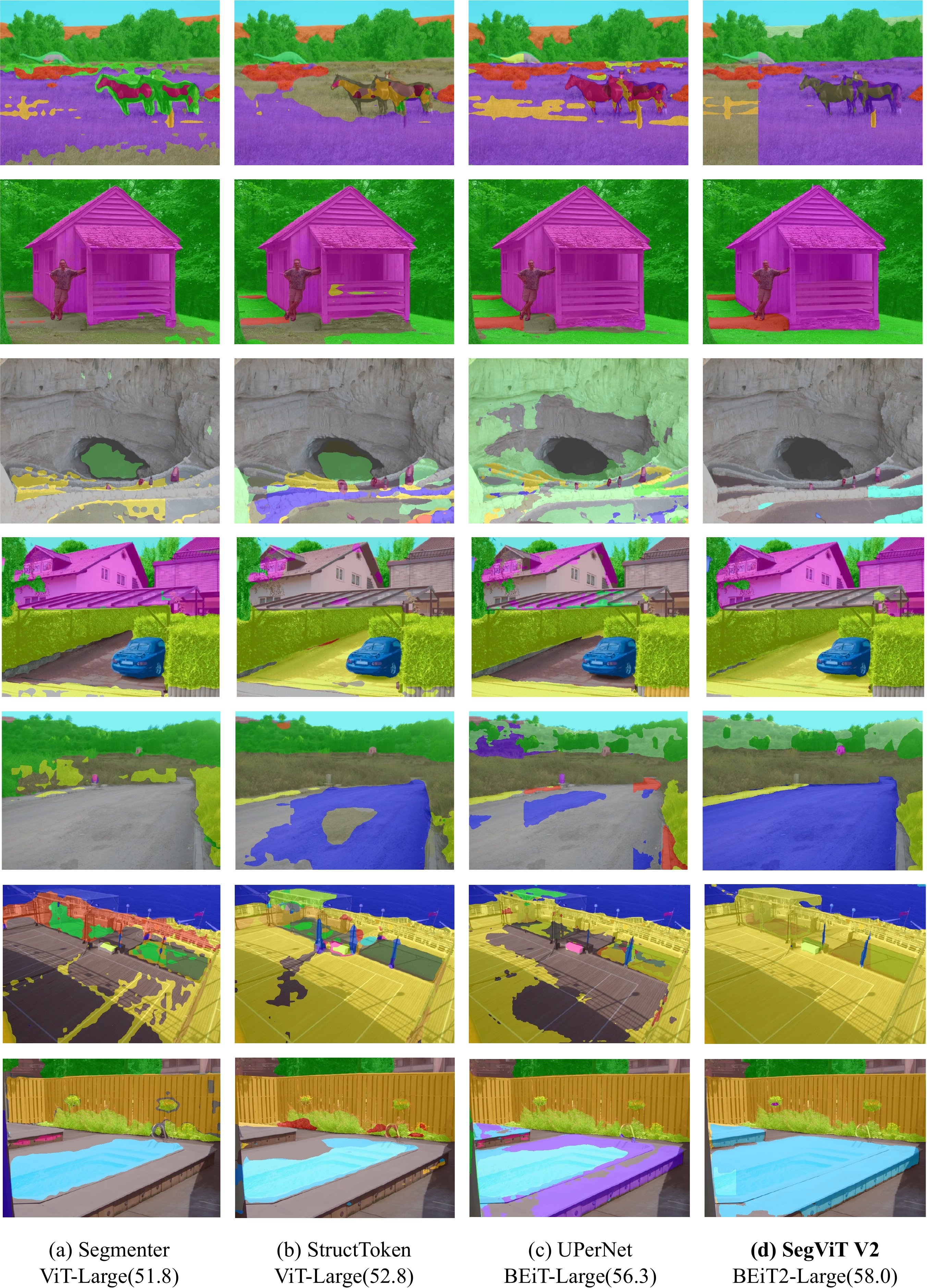}
    \caption{ Visuals results of different segmentation networks and plain ViT backbones on the ADE20K validation set~\cite{ade20k}. It includes the following models: (a) Segmenter \cite{strudel2021segmenter} with ViT large, (b) StructToken \cite{lin2022structtoken} with ViT large, (c) UPerNet \cite{Upernet} with BEiT large, and (d) SegViT V2 with BEiTv2 large. The results demonstrate that our methods effectively generate accurate segmentation masks and unlock the potential of plain ViT. Zoom in for a better view. }
    \label{fig:viz_comapre}
\end{figure*}

\subsection{Comparisons with the State-of-the-art Methods}
\paragraph{Results on ADE20K.}
\cref{tab:ade20k} reports the comparison with the state-of-the-art methods on ADE20K validation set using ViT backbone. The \seg\ uses the \atm\ module with multi-layer inputs from the original ViT backbone, while the \emph{Shrunk} is the one that conducts QD to the ViT backbone and  saves $50\%$ of the computational cost without sacrificing too much performance. Our approach achieves a state-of-the-art mIoU of $58.2\%$ (MS) with the BEiTv2 Large backbone.
To ensure a fair comparison, we evaluate our \seg\ module with the BEiT-v2 large backbone on a crop size of $512\times512$, which consumes 374.0 GFlOPs. Our approach achieves a slightly better performance of $56.5\%$ mIoU compared to Mask2former-Swin-L, which achieves $56.1\%$ with 402.7 GFlops on a crop size of $640\times640$. Additionally, our \emph{Shrunk} version offers around a 50\% reduction in computational cost (308.8 GFLOPs), while delivering competitive performance with a mIoU of $57.0\%$ (MS). Optimizing SegViT with ViT-Large using the proposed \emph{Shrunk++} reduces the computational cost of \emph{Shrunk} by 3.05 times, while preserving the mIoU. 
\cref{fig:viz_comapre} shows the visual results of different segmentation methods. In contrast to other methods that often confuse similar classes and misclassify related concepts, our SegViT stands out by more precise object boundary delineation and achieving accurate segmentation of complete objects, even in cluttered scenes. %

\paragraph{Results on COCO-Stuff-10K.}
\cref{tab:cocostuff} shows the result on the COCO-Stuff-10K dataset. Our method achieves $50.3\%$ which is higher than the previous state-to-the-art StrucToken by $1.2\%$ with less computational cost. Our \emph{Shrunk} version achieves $49.4\%$ mIoU with $224.8 $ GFLOPs, which is similar to the computational cost of a dilated ResNet-101 backbone but with much higher performance. By extending SegViT with the effective \emph{Shrunk++}, we significantly decrease its GFLOPs by 1.82 times, while retaining a competitive mIoU.

\paragraph{Results on PASCAL-Context.}
\cref{tab:pascal} shows the results on the PASCAL-Context dataset. We follow HRNet~\cite{hrnet} to evaluate our method and report the results  under $59$ classes (without background) and $60$ classes (with background). 
Using full SegViT structure without adopting \emph{Shrunk} or \emph{Shrunk++}, we reach mIoU of $67.14\%$ and $61.63\%$ respectively for those two metrics, outperforming the state-of-the-art methods using the ViT backbones with less computational cost. By applying \emph{Shrunk} and \emph{Shrunk++} architecture, the computational cost in terms of GLOPs is reduced by $42\%$ and $45\%$, respectively. Among all approaches evaluated on the PASCAL-Context dataset, SegViTv2 with \emph{Shrunk++} achieves the best trade-off between accuracy and efficiency.

\subsection{Ablation Study}
In this section, we conduct extensive ablation studies to show the effectiveness of our proposed methods.

\paragraph{Effect of the ATM module.} We conducted an analysis to evaluate the impact of using the proposed ATM module as an encoder. The results are summarized in Table~\ref{tab:mask_pred}. To establish a baseline for comparison, we introduced SETR-naive, which utilizes two $1\times 1$ convolutions to directly derive per-pixel classifications from the final layer of the ViT-Base transformer output.
From the results, it is evident that applying the ATM module under the supervision of a conventional cross-entropy loss leads to a performance improvement of 0.5\%. However, the performance gains become much more substantial when we decouple the classification and mask prediction processes, supervising each separately. This approach results in a significant performance boost of 3.1\%, highlighting the efficacy of the ATM module in enhancing semantic segmentation performance.

\begin{table}[!ht]
    \centering
        \caption{
        Comparisons between our proposed \atm\ module with SETR~\cite{setr}. `CE loss' indicates the cross-entropy loss commonly used in semantic segmentation. The experiments on the ADE20k dataset are carried out using the ViT-Base backbone. 
        }
        \vspace{0.5em}
    \label{tab:mask_pred}
    \begin{tabular}{rll }
    \toprule
    Decoder    & Loss & mIoU (ss) \\
    \midrule
     SETR & CE loss &   46.5\\
     \atm  &  CE loss&  47.0 (+0.5)\\
     \atm  &   $\mathcal{L}_{mask}$ loss&  \textbf{49.6 (+3.1)} \\

    \bottomrule
    \end{tabular}
\end{table}

\paragraph{Ablation of 
the feature levels.}
The effects of using multiple-layer inputs from the backbone to the ATM modules are presented in Table~\ref{tab:atm_layer}. The incorporation of feature maps from lower layers leads to a notable performance improvement of 1.3\%. We further investigated the impact of including more layers of features and observed additional gains in performance. After empirical testing, we determined that utilizing three layers yielded optimal results, resulting in an overall mIoU boost of 1.7\%.
These ablation studies confirm the effectiveness of our proposed ATM decoder and highlight the advantage of incorporating multi-layer features into the segmentation structure. This integration significantly enhances the performance of semantic segmentation tasks.

\begin{table}[!ht]
    \centering
        \caption{Results of using different layer inputs to the \seg\ structure on ADE20K dataset using ViT-Base as the backbone. Involving multi-layer features leads to obvious performance gains.}
        \vspace{.5em}
            \label{tab:atm_layer}
    \begin{tabular}{lrl}
    \toprule
        &Used layers & mIoU (ss) \\
        \midrule
        Single layer & {[12]} &  49.6  \\
        Cascade &{[6, 12] }&  50.9 (+1.3)\\
        Cascade &{[6, 8, 12] }& \textbf{51.3 (+1.7)} \\
        Cascade &{[3, 6, 9, 12] }& 51.2 (+1.6) \\
    \bottomrule
    \end{tabular}
\end{table}

\paragraph{\seg\ on hierarchical base models.} We conducted an analysis to evaluate the performance of SegViT on hierarchical base models. For comparison, we selected two competitive methods, Maskformer~\cite{maskformer} and Mask2former~\cite{cheng2021mask2former}. The results presented in Table~\ref{tab:hierarchi} indicate that, even though our method was not specifically designed for hierarchical base models, we are still able to achieve competitive performance while maintaining computational efficiency. This demonstrates the applicability of our SegViT approach to various types of ViT-Base models.

 \begin{table}[!ht]
  \centering
      \caption{The experiments use the Swin-Tiny \cite{liu2021swin} backbone and are carried out on the ADE20K dataset. The GFLOPs are measured at single-scale inference with a crop size of $512 \times 512$.  
       }
    \vspace{0.5em}
     \label{tab:hierarchi}
    \begin{tabular}{rcc}
   \toprule
        Method & mIoU (ss) & GFLOPs \\
       \midrule
       Maskformer \cite{maskformer} & 46.7 & 57.3\\
        Mask2former \cite{cheng2021mask2former} & \textbf{47.7} & 73.7 \\
       \seg\ (Ours) & 47.1 & \textbf{48.0}\\
        \bottomrule
    \end{tabular}
\end{table}

 \begin{table*}
	\centering
    \vspace{-1em}
        \caption{Ablation results of \emph{Shrunk} and \emph{Shrunk++} version on the ADE20K dataset. We explored various shrink strategies. The GFLOPs are measured at single-scale inference with a crop size of $512 \times 512$ on the ViT-Base backbone. QD: query-based downsampling. QU: query-based upsampling. $QD_{\text{layer}}$  indicates which layer to apply the QD. $QD_{\text{method}}$ indicates the downsampling method for QD.
        }
            \label{tab:tpn ablation}
    \begin{tabular}{lccccllc}
    \toprule
        Structure & QD & QU & $QD_{\text{layer}}$ &$QD_{\text{method}}$& Head & mIoU (ss) & GFLOPs \\
        \midrule
        Single & - & -&-&-& SETR & 46.5 & 107.3 \\
        Single  & - & -&-&-& \atm & 49.6 (+3.1) & 115.8 \\
        Naive \emph{Shrunk} & \checkmark &-&6&2x2 & \atm & 46.9 (+0.4) & 74.1 \\
        \emph{Shrunk} & \checkmark &\checkmark & 6&2x2& \atm & \textbf{50.0 (+3.5)} & 97.1\\
        \midrule
        Nearest - TS & \checkmark &\checkmark & 0&3x3& \atm & 38.9(-7.6) & 32.8\\
        Nearest - TS & \checkmark &\checkmark & 0&2x2& \atm & 43.3(-3.2) & 46.1\\
        Shrunk++ & \checkmark &\checkmark& 0 &3x3-Edge& \atm & 47.9(+1.4) & 69.3 \\
        Shrunk++ & \checkmark &\checkmark& 0 &2x2-Edge& \atm & \textbf{49.9(+3.4)} & \textbf{74.6}\\

    \bottomrule
    \end{tabular}
\end{table*}

\begin{table*}[bp]
\parbox{.65\linewidth}{
\centering
\setlength{\tabcolsep}{1.5pt}
\footnotesize
\caption{Ablation results of different decoder methods with their corresponding feature merge types and loss types. ViT-Base is employed as the backbone for all the variants.}
\vspace{0.5em}
\label{tab:struct compare}

\resizebox{1.0\linewidth}{!}{\begin{tabular}{rcccccc}
\toprule
\multicolumn{1}{c}{} & \multicolumn{2}{c}{Multi-level Features} & \multicolumn{3}{c}{Loss Types} & \multicolumn{1}{c}{} \\ 
\cmidrule(r){2-3}
\cmidrule(l){4-6}
\multicolumn{1}{c}{Decoder} & FPN & \multicolumn{1}{c}{Token Merge} & Pixel level & Dot product & Attention Mask & \multicolumn{1}{c}{mIoU (ss)} \\ 
\midrule
SETR-MLA \cite{setr}& \checkmark & \multicolumn{1}{c}{} & \checkmark &  &  & 48.2 \\
Segmenter \cite{strudel2021segmenter} &  & \multicolumn{1}{c}{} & \checkmark &  &  & 49.0 \\
MaskFormer \cite{maskformer} & \checkmark & \multicolumn{1}{c}{} &  & \checkmark &  & 46.7 \\
Ours-Variant 1 &  & \multicolumn{1}{c}{} &  &  & \checkmark & 49.6 \\
Ours-Variant 2&  & \multicolumn{1}{c}{\checkmark} &  & \checkmark &  & 50.6 \\
Ours &  & \multicolumn{1}{c}{\checkmark} &  &  & \checkmark & 51.2 \\ 
\bottomrule
\end{tabular}}}
\hfill
\parbox{.3\linewidth}{
    \centering
        \caption{Ablation of the QD module in terms of the targets and methods to down-sample. The experiments are carried out on the ViT-Large backbone of ADE20K dataset. }
        \vspace{.5em}
            \label{tab:downsample}
    \begin{tabular}{lcc}
    \toprule
    Applied to & Methods &  mIoU (ss)  \\
    \midrule
        Q & Conv & 44.5  \\
        Q, K, V & Nearest &  52.6  \\
        Q & Nearest & \textbf{53.9} \\
    \bottomrule
    \end{tabular}
}
\end{table*}

\paragraph{Ablation of \emph{Shrunk} and \emph{Shrunk++} strategies.} In this section, we analyze the effectiveness of the different SegViT structures. Table~\ref{tab:tpn ablation} presents the effects of various techniques employed in each SegViT structure, including query upsampling (QU), query downsampling (QD), token-squeezing (TS) techniques, and segmentation heads.
Applying the ATM head to the 'Single' structure yields a notable performance improvement of 6.67\% compared with using the SETR head.  This demonstrates the effectiveness of the ATM head in enhancing the performance of the baseline structure.
However, applying QD to the 'Single' structure with the ATM head leads to a performance drop of 2.7\%, suggesting the occurrence of information loss during the downsampling phase.
Importantly, incorporating QU restores the performance. QU helps recover the discarded information from QD and reconstructs the high-resolution feature map, which is crucial for dense prediction tasks.
Jointly leveraging QU and QD, the \emph{Shrunk} architecture achieves optimal performance while reducing computational costs by 16.15\% in comparison to the `Single' structure.

In the proposed \emph{Shrunk++} structure, we analyze the performance of two main token-squeezing techniques: nearest downsampling and edge-aware downsampling. It is important to note that token squeezing is directly applied to the first layer of the network for optimal computational efficiency.
Applying naive nearest downsampling with a 3x3 kernel reduces the GFLOPs of the \emph{Shrunk} structure without token-squeezing by a factor of 2.97. However, reducing the computational cost with 3x3 and 2x2 nearest downsampling leads to a performance drop of 13\%.
In contrast, by incorporating an additional edge extractor into our \emph{Shrunk++} architecture, we significantly improve the mIoU, achieving performance on par with \emph{Shrunk}, i.e., 49.9\% mIoU, with a minor increase in computational cost to 74.6 GFLOPs. The edge-aware downsampling technique preserves the edge details, thereby preserving discriminative features for dense predictions.
Among the different settings, the 2x2 + Naive MLP Edge setting achieves an optimal balance between performance and efficiency.

\paragraph{Ablation studies on decoder variances.}
Different decoder methods are associated with specific feature merge types and loss types. In Table~\ref{tab:struct compare}, we compare the designs of various decoders on a plain ViT backbone.
For hierarchical base models like Swin, the resolution of the feature maps in each stage is reduced. Consequently, the adoption of a Feature Pyramid Network (FPN) is necessary to obtain feature maps with larger resolutions and rich semantic information. However, in Table~\ref{tab:struct compare}, we observe that the FPN structure does not perform well with plain vision transformers.
With plain ViT base models, the resolution remains constant, and the feature map of the final layer encapsulates the most comprehensive semantic information.
Hence, our proposed method, which utilizes tokens to merge features from different levels, achieves superior performance. By simply replacing the FPN structure with the ATM-based token merge, we improve the performance from 46.7\% to 50.6\%.
Regarding the loss type, the pixel-level loss refers to the conventional cross-entropy loss applied to the feature map. The dot product loss corresponds to the loss utilized in \cite{detr} and \cite{maskformer}.  
Attention mask loss indicates the direct application of mask supervision to the similarity map generated by the ATM during attention calculation. 
Incorporating loss supervision on the attention mask, as in our method, leads to a performance improvement of 0.6\%.

\newcommand{\tri}{\begin{small}\color{Green}$\blacktriangle$\end{small}}
\newcommand{\trid}{\begin{small}\color{BrickRed}$\blacktriangledown$\end{small}}
\newcommand{\trim}{\begin{small}\color{SpringGreen}$\blacktriangledown$\end{small}}

\begin{table*}[!t]
\centering
\caption{
Comparisons for various ViT pre-training schedules on the validation set of ADE20K. All results are reported in single-scale inference. The default configuration for these base models is pre-trained on ImageNet-1K with 224 * 224 resolutions. `*' means the models use the backbone weights pre-trained with 384 * 384 resolutions. '\textsuperscript{\textdagger}' means the base models pre-trained on imagenet-21K. The proposed SegVit head has a less computational cost and performs better than UPerNet among all pre-training variants.}
\label{tab:backbone}
\begin{tabular}{rccccc} 
\toprule
Backbone              & SegViT mIoU  & Head FLOPs    & UPerNet mIoU & Head FLOPs  & ImageNet Acc          \\ 
\midrule
MAE Base~\cite{he2022masked}      & 49.22 (\tri1.12) & 6.89(\trid329.73) & 48.1  & 336.62 & 83.66  \\
CLIP Base~\cite{radford2021learning}             & 50.76 (\tri1.16) &  6.89(\trid329.73)  & 49.6 & 336.62  & 80.20    \\
CAE Base~\cite{chen2022context}     & 50.42 (\tri0.22) &  6.89(\trid329.73)  & 50.2 & 336.62  & 83.90     \\
iBot Base~\cite{zhou2021ibot}         & 50.58 (\tri0.58) & 6.89(\trid329.73)  & 50.0 & 336.62  & 84.00            \\

Augreg Base*\textsuperscript{\textdagger}\cite{augreg}  & 51.30 (\tri2.66) & 6.89(\trid329.73)  & 48.6  & 336.62  & 85.49 \\
DEiT v3 Base\textsuperscript{\textdagger}\cite{touvron2022deit}  & 52.40 (\tri0.60) & 6.89(\trid329.73) & 51.8 & 336.62  & 85.70       \\
BEiT v2 Base\textsuperscript{\textdagger}\cite{beitv2}  & 53.97 (\tri0.47) & 6.89(\trid329.73) & 53.5   &  336.62 & 86.50       \\ 
\hline\hline
Augreg Large*\textsuperscript{\textdagger}\cite{augreg} & 54.60 (\tri2.50)        & 16.36(\trid1,366.33) & 52.1 & 1382.69  & 85.59           \\
DEiT v3 Large*\textsuperscript{\textdagger}\cite{touvron2022deit} & 55.81 (\tri1.21)          & 16.36(\trid1,366.33)   & 54.6 & 1382.69  & 87.70            \\
BEiT v2 Large\textsuperscript{\textdagger}\cite{beitv2}  &    58.00 (\tri1.30)         & 16.36(\trid868.28) & 56.7  & 884.64 & 87.30            \\
\bottomrule
\end{tabular}
\end{table*}

\paragraph{Ablation for the QD module.}
The motivation behind using QD is to leverage the pre-trained weights of the backbone. 
As shown in Table~\ref{tab:downsample}, using a stride-2 convolution with learnable parameters to downsample the query will disturb the pre-trained weights, leading to a notable decline in performance.
Applying down-sampling to both the query and the key-value pairs would inevitably lead to information loss during the down-sampling process, which is evident in the lower performance.
Our results show that 
applying $2 \times 2$ nearest down-sampling exclusively to the query in the QD module yields better results. This approach allows us to preserve the pre-trained weights of the backbone while achieving the desired down-sampling effect.

\subsection{Application 1: A Better Indicator for Feature Representation Learning} 
\paragraph{Background.} Semantic segmentation serves as a fundamental vision task that has been extensively employed in previous research to assess the representation learning capabilities of weakly, fully, and self-supervised base models~\cite{he2022masked,chen2022context,touvron2022deit,beitv2}. In prior work, the UPerNet decoder structure has been commonly used for semantic segmentation. However, the UPerNet decoder may not be a suitable indicator for evaluating the feature representation ability of the base model. This is primarily due to its heavier computational requirements and slower convergence rate. 
Additionally, variations in feature representation acquired by the base model can be substantial due to diverse training strategies during the fine-tuning process on semantic segmentation datasets
Consequently, the task of semantic segmentation may not adequately evaluate the feature representation ability of pre-trained models.

\paragraph{Experiment settings.} 
In this section, we extensively evaluate our proposed SegVit across diverse weakly, fully, and self-supervised vision transformers, 
including those proposed by He et al.~\cite{he2022masked}, Chen et al.~\cite{chen2022context}, Touvron et al.~\cite{touvron2022deit}, and the BEiT model~\cite{beitv2}. We demonstrate that our method outperforms UPerNet~\cite{Upernet} in both self-supervised and multi-modality base models, achieving state-of-the-art performance. Notably, our approach achieves superior performance to UPerNet while utilizing only 5\% of the computational cost in terms of the decoder head.
Table~\ref{tab:backbone} illustrates that our proposed SegViT head consistently outperforms UPerNet across all base models. For the ViT-Base, our method improves the performance of UPerNet on the CLIP model by 1.16\% while significantly reducing the computational cost. Similar findings are evident 
for ViT-Large base models. Furthermore, compared to UPerNet, our proposed SegViT's decoder head exhibits a better alignment between the growth trend of segmentation accuracy and the classification accuracy on ImageNet. This clearly demonstrates the superior efficiency of our SegViT head compared to UPerNet, making it a more suitable indicator for representation learning in base models.

\newcommand{\trian}{\begin{small}\color{Black}$\blacktriangledown$\end{small}}

\begin{table*}[!t]
\centering
\caption{CSS results on ADE20k in mIoU (\%) on 100-50 and 100-10 settings. 
The relative mIoU reduction compared with the joint training for each method is reported.
}
\label{tab:ade_sota}
\resizebox{1.0\textwidth}{!}{%
\begin{tabular}{l|cccc||cccc@{}}
\toprule
& \multicolumn{4}{c}{\textbf{100-50} (2 tasks)} & \multicolumn{4}{c}{\textbf{100-10} (6 tasks)} \\
\textbf{Method} & 0-100 & 101-150 & \textit{all} & \textit{avg} & 0-100 & 101-150 & \textit{all} & \textit{avg}\\
\midrule
ILT~\cite{Michieli2019IncrementalLT}& 18.29 (\trian 26.1)  & 14.40 (\trian 13.8)  & 17.00 (\trian 22.0)  & 29.42  & 0.11 (\trian 44.2)  &  3.06 (\trian 25.1) & 1.09 (\trian 37.9) & 12.56 \\ 

MiB~\cite{cermelli2020ModelingTB}& 40.52 (\trian3.9)  & 17.17 (\trian11.0)  & 32.79 (\trian6.2)  & 37.31 &  38.21 (\trian6.1) & 11.12 (\trian 17.1) & 29.24 (\trian 9.8) & 35.12 \\

SDR~\cite{michieli2021ContinualSS} & 40.52 (\trian 3.8)  & 17.17 (\trian 11.0)  & 32.79 (\trian 6.2)  & 37.31  & 37.26 (\trian 7.1) & 12.13 (\trian 16.1)  & 28.94 (\trian 10.1) & 34.48 \\

PLOP~\cite{douillard2021PLOPLW} & 41.76 (\trian 2.6) & 14.52 (\trian 13.7) & 32.74 (\trian 6.3) & 37.73 & 38.59 (\trian 5.8) & 14.21 (\trian 14.0) & 30.52 (\trian 8.5) & 34.48 \\

REMINDER~\cite{phan2022class} & 41.55 (\trian 2.8) & 19.16 (\trian 9.0)  & 34.14 (\trian 4.9)  & 38.43 & 38.96 (\trian 5.4)  & 21.28 (\trian 6.9) & 33.11 (\trian 5.9) & 37.47 \\

RCIL~\cite{zhang2022representation}  &  42.35 (\trian 2.0)	&  18.47 (\trian 9.7) & 34.45 (\trian 4.6) & 38.48 &  29.42 (\trian 15.0)	& 13.49 (\trian 14.0)	 & 28.36 (\trian 10.0) & 29.93 \\

Oracle - ResNet backbone & 44.34 & 28.21 &39.00 & - & 44.34 & 28.21 &39.00 & - \\
\midrule 

MiB~\cite{cermelli2020ModelingTB} & 43.43 (\trian 3.2)  & 30.63 (\trian 4.3)  & 39.19 (\trian 3.6) & 38.66 & 39.15 (\trian 7.5) & 20.37 (\trian 14.5) & 34.17 (\trian 8.6) & 39.53 \\

PLOP~\cite{douillard2021PLOPLW} & 43.82 (\trian 2.8) & 26.23 (\trian 8.7)  & 37.99 (\trian 4.8) & 38.06 & 43.25 (\trian 3.4) & 24.13 (\trian 10.8) & 36.25 (\trian 6.5) & 40.28 \\

REMINDER~\cite{phan2022class} & 44.66 (\trian 2.0) & 26.76 (\trian 8.1)  & 38.73 (\trian 4.0)  & 38.43 & 43.28 (\trian 3.4)  & 24.33 (\trian 10.6) & 37.10 (\trian 5.6) & 41.76 \\

Oracle - ViT backbone & 46.63 & 34.90 & 42.75 & - & 46.63 & 34.90 & 42.75 & - \\\midrule 

SegViT-CL (ours) & \textbf{53.64 (\trian 0.5)}    & \textbf{40.00 (\trian 5.6)}   & \textbf{49.09 (\trian 2.2)} & 46.82 & \textbf{53.77 (\trian 0.3)}          & \textbf{35.54 (\trian 10.0)}        & \textbf{47.70 (\trian 3.6)} & 50.59 \\
Oracle & 54.11 & 45.60 & 51.28       &  -    & 54.11 & 45.60 & 51.28        &  -    \\ 
\bottomrule
\end{tabular}%
}
\end{table*}

\subsection{Application 2: Continual Semantic Segmentation}
The decoupling of class prediction and mask segmentation in our proposed SegVit decoder makes it inherently well-suited for continual learning settings. This characteristic allows us to learn new classes by solely fine-tuning the class proxy (the class token), leveraging the powerful representation ability of the plain vision transformer while keeping the old parameters frozen. To validate the effectiveness of this new approach to continual learning, we conducted experiments following standard settings adopted by prior studies.

\paragraph{Experiment settings.} Continual Semantic Segmentation (CSS) has two settings~\cite{cermelli2020ModelingTB,douillard2021PLOPLW}: disjoint and overlapped.
In the disjoint setup, all pixels in the images at each step belong to the previous classes or the current class.
In the overlapped setting, the dataset of each step contains all the images that have pixels of at least one current class, and all pixels from previous and future tasks are labeled as background.
The overlapped setting is more realistic and challenging, thus we evaluate the performance of the overlapped setup on the ADE20k dataset.

Following prior studies~\cite{phan2022class,cermelli2020ModelingTB,douillard2021PLOPLW}, we perform three experiments: adding 50 classes after training with 100 classes (100-50 setting with 2 steps), adding 50 classes each time after training with 50 classes (50-50 setting with 3 steps), adding 10 classes each time sequentially after training with 100 classes (100-10 setting with 6 steps).

\paragraph{Baselines} We conducted a comprehensive comparison of our proposed method against state-of-the-art Continual Semantic Segmentation (CSS) techniques, including RCIL~\cite{zhang2022representation}, PLOP~\cite{douillard2021PLOPLW}, REMINDER~\cite{phan2022class}, SDR~\cite{michieli2021ContinualSS}, and MiB~\cite{cermelli2020ModelingTB}.
To ensure fair comparisons, existing methods were evaluated using DeepLabV3~\cite{chen2017rethinking} with ResNet101 and ViT-Base backbones that were pre-trained on ImageNet-21k. The reported results for PLOP, RCIL, and REMINDER were obtained based on the codebases provided by the respective authors. Furthermore, we included the performance of the Oracle model, which represents the upper bound achieved by jointly training on all available data, serving as a benchmark for each method.

\begin{figure}[t]
    \centering
    \includegraphics[width=\linewidth]{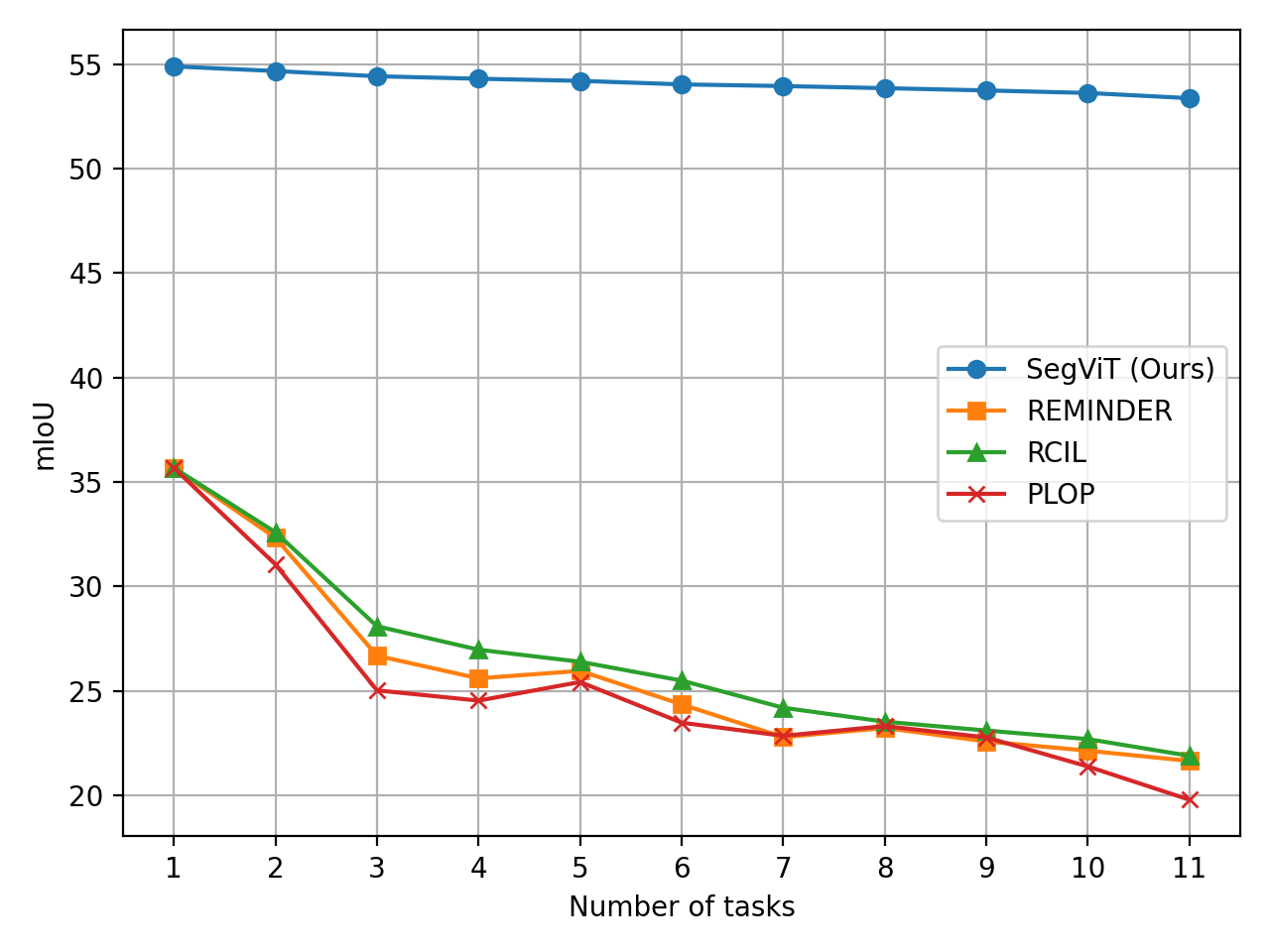}
    \caption{mIoU of recent CSS methods on the first 100 base classes after incrementally learning new tasks on 100-5 settings with 11 tasks.}
    \label{fig:css}
\end{figure}

\textbf{Metrics.} We evaluate the model performance by five mIoU metrics.
First, we compute mIoU for the base classes $\cC^0$, which reflects model rigidity: the model's resilience to catastrophic forgetting.
Second, we compute mIoU for all incremented classes $\cC^{1:T}$, which measures plasticity: the model capacity in learning new tasks.
Third, we compute the mIoU of all classes in $\cC^{0:T}$ (\emph{all}), which shows the overall performance of models. 
Fourth, we report the average of mIoU (\emph{avg}) measured step after step as proposed by~\cite{douillard2021PLOPLW}, which evaluates performance over the entire continual learning process. To ensure fair comparisons, we evaluate the relative performance of each CSS method in terms of relative mIoU reduction compared with its Oracle model, jointly trained on all data. %

\begin{table}[t]
\centering
\caption{Performance drop (degree of forgetting) of all classes grouped by tasks on the 100-10 setting. We report the class mIoU when the model first learns the task, and the mIoU when the model last learns it.}
\label{tab:task_drop}
\resizebox{0.48\textwidth}{!}{%
\begin{tabular}{c|cccccc} 
\toprule
Tasks            & 101-110 & 111-120 & 121-130 & 131-140 & 141-150 & avg \\ 
\midrule
First Time & 34.93   & 39.78  &  41.10  & 36.22   & 27.95 & 35.99 \\
Last Time  & 34.51   &  39.30  & 40.86   & 35.09  & 27.95 &   35.54\\ 
\midrule
Forgetting & \trian 0.42 & \trian 0.48 & \trian 0.24 & \trian 1.12 & \trian 0 & \trian 0.45 \\
\bottomrule
\end{tabular}%
}
\end{table}

\paragraph{Results and Discussion.}
Table~\ref{tab:ade_sota} shows the results of different CSS methods on ADE20k. Our SegViT-CL consistently outperforms existing methods in \emph{all} mIoU for both settings. In terms of mIoU reduction, the proposed SegViT-CL only decreases the mIoU of the Oracle model by $2.2\%$ on the 100-50 setting, which is two times better than the second-best method, RCIL with ResNet backbone with $4.6\%$ reduction. This substantial enhancement over existing methods underlines the effectiveness of our proposed method in the continual semantic segmentation paradigm. On a long CL setting 100-10 with 6 tasks, ours is almost forgetting-free with a marginal mIoU reduction of $0.3\%$, while recent CSS methods significantly suffer from forgetting with at least $5.4\%$ mIoU reduction. Using the ViT backbone, existing methods including MiB, REMINDER, and PLOP still suffer from high mIoU reductions. Compared with the Oracle, MiB~\cite{cermelli2020ModelingTB}, PLOP~\cite{douillard2021PLOPLW}, and REMINDER~\cite{phan2022class} decrease the mIoU by 8.6\%, 6.5\% and 5.6\% respectively on the 100-10 setting, demonstrating the sub-optimal performance of current CSS methods for ViT architecture. This highlights the need for developing a specialized ViT architecture that is robust to forgetting.

To evaluate the forgetting of every task on the 100-10 setting, we compute the performance drop at the last step compared with its initial mIoU when the model first learns the task. For example, the initial mIoU of task 2 is the mIoU of class 101-110 evaluated at step 2. Similarly, that of task 3 is the mIoU of class 111-120 reported at step 3. \cref{tab:task_drop} shows the performance drop at the last step compared with the initial mIoU of each task. Averaged across 5 tasks, the mIoU only drops by 0.45\%, which shows that SegViT is robust to forgetting across all tasks on the 100-10 setting.  \cref{fig:css} shows the mIoU on the base classes after incrementally training on many tasks in 100-5, which is a long continual learning setting with 11 tasks. Overall, our SegViT achieves nearly zero forgetting for almost all tasks at the last step. %
 In contrast to previous CSS methods which require partial fine-tuning, the proposed SegViT supports completely freezing old parameters, effectively eliminating any interference with previously acquired knowledge.

\section{Conclusion}
This paper presents SegViTv2, a novel approach for semantic segmentation using plain ViT transformer base models. The proposed method introduces a lightweight decoder head that incorporates the Attention-to-mask (\atm) module. Additionally, a \emph{Shrunk++} structure is proposed to reduce the computational cost of the ViT encoder by 50\% while maintaining competitive segmentation accuracy. 
Moreover, this work extends the SegViT framework to address the challenge of continual semantic segmentation, aiming to achieve nearly zero forgetting. By protecting the parameters of old tasks, SegViT effectively mitigates the impact of catastrophic forgetting. 
Extensive experimental evaluations conducted on various benchmarks demonstrate the superiority of SegViT over UPerNet, while significantly reducing computational costs. The introduced decoder head provides a robust and cost-effective avenue for future research in the field of ViT-based semantic segmentation.

\section*{Acknowledgments}
 This work was in part supported by the National Key R\&D Program of China (No.\  2022ZD0118700).
  Y. Liu's participation was in part supported by the start-up funding of  The University of Adelaide.  
  We express our gratitude to The University of Adelaide High-Performance Computing Services for providing the GPU Compute Resources, and to Mr. Wang Hui and Dr. Fabien Voisin for their valuable technical support for the training infrastructure.

\bibliographystyle{IEEEtran}
\bibliography{main}

\end{document}